# Probabilistic automata for computing with words

Yongzhi Cao[a,*,1], Lirong Xia[a], Mingsheng Ying[a,2]

[a]*State Key Laboratory of Intelligent Technology and Systems, Department of Computer Science and Technology, Tsinghua University, Beijing 100084, China*

**Abstract**

Usually, probabilistic automata and probabilistic grammars have crisp symbols as inputs, which can be viewed as the formal models of computing with values. In this paper, we first introduce probabilistic automata and probabilistic grammars for computing with (some special) words in a probabilistic framework, where the words are interpreted as probabilistic distributions or possibility distributions over a set of crisp symbols. By probabilistic conditioning, we then establish a retraction principle from computing with words to computing with values for handling crisp inputs and a generalized extension principle from computing with words to computing with all words for handling arbitrary inputs. These principles show that computing with values and computing with all words can be respectively implemented by computing with some special words. To compare the transition probabilities of two near inputs, we also examine some analytical properties of the transition probability functions of generalized extensions. Moreover, the retractions and the generalized extensions are shown to be equivalence-preserving. Finally, we clarify some relationships among the retractions, the generalized extensions, and the extensions studied recently by Qiu and Wang.

## 1  Introduction

To capture the notion of automated reasoning involving linguistic terms rather than numerical quantities, Zadeh has proposed and advocated the idea of computing with words in a series of papers [23, 25, 26, 27, 28, 29]. The objects of computing with words are words and propositions which describe perceptions in a natural language, where the words play the role of labels of perceptions. For the purpose of computing, the meaning of a proposition is expressed as a generalized constraint. Many basic types of constraints have already been given by Zadeh; among others, possibilistic constraint characterized by fuzzy sets (possibility distributions) and probabilistic constraint characterized by probabilistic distributions are two most familiar ones. As a methodology, computing with words has provided a foundation for dealing with imprecise, uncertain, and partially

* Corresponding author.

*E-mail addresses:* caoyz@mail.tsinghua.edu.cn (Y.Z. Cao), xialirong@gmail.com(L. Xia), yingmsh@mail.tsinghua.edu.cn (M.S. Ying).

[1] Supported in part by the National Foundation of Natural Sciences of China under Grants 60496321, 60321002, and 60505011.

[2] Supported in part by the Key Grant Project of Chinese Ministry of Education under Grant 10403.

true data which have the form of propositions expressed in a natural language; see [7, 9, 21, 30] for some applications.

Based upon the generalized constraints, one can manually handle some computations and uncertain reasoning on perceptions. However, computing, in its traditional sense, is centered on the manipulation of numbers and symbols, and is usually represented by a dynamic model in which an input device is equipped. It is well known that various automata, such as Turing machines, deterministic and nondeterministic finite automata, probabilistic automata, and fuzzy automata, are the prime examples of classical computational systems. Note that the inputs of such models are exact rather than vague data, and thus they cannot serve as formal models of computing with words. This observation motivates Ying to interpret "computing with words" as a computational procedure where inputs are allowed to be vague data [22]. In this sense, it is, however, not easy to implement the computations on perceptions, because we have not known what formal models are competent for these computations. Of course, this may be alleviated by providing more candidate models. But even if a formal model is picked out, how to select some words as inputs remains difficult since usually the designer only provides specification for finite words. On the other hand, if a word $W$ is selected as an input, then a word $W'$ near to $W$ should also be selected because the description of a perception in a natural language is generally not precise and we have no excuse for rejecting a similar label of the perception as an input. This consideration puts us in a dilemma since allowing similar words as inputs will lead to an infinite input alphabet.

Most of the literature on computing with words is devoted to developing new computationally feasible algorithms for uncertain reasoning; however, to our knowledge, few efforts have been made to consider the formal theory of computing with words except the work [22, 19, 12]. In [22], Ying proposed a formal model of computing with words in terms of fuzzy automata. Fuzzy automata initiated by Santos [15] are a generalization of nondeterministic finite automata, in which state transitions are imprecise and uncertain. The point of departure in [22] is a fuzzy automaton where inputs are crisp symbols. These symbols may be reasonably thought of as the input values that we are going to compute. Following [22], we identify a value with a symbol from the input alphabet and also a word with a probabilistic distribution or a fuzzy subset of the input alphabet, and use them exchangeably. By exploiting Zadeh's extension principle, the fuzzy automaton gives rise to another fuzzy automaton that has all fuzzy subsets of the set of the symbols as inputs and models formally computing with words. The key idea underlying Ying's formal model of computing with words is the use of words in place of values as input symbols of a fuzzy automaton. Motivated by this, Wang and Qiu extended the concept of computing with words to fuzzy Turing machines and some formal grammars in [19]; moreover, they investigated the formal theory of computing with words in the framework of probabilistic automata and probabilistic grammars [12], where the words are interpreted as probabilistic distributions.

Essentially, the buildings of all the formal models of computing with words in [22, 19, 12] go as follows: beginning with a classical computational model with values as inputs and then deriving a formal model with all words (interpreted as probabilistic distributions or possibility distributions) as inputs. Consequently, the resultant formal model for computing with words inevitably depends on the underlying classical computational model of computing with values. This observation suggests us seek a general formal model of computing with words. At this point, we introduced the notion of fuzzy automata for computing with words (FACWs) in [2]. An FACW is a fuzzy automaton where the input alphabet consists of finite words (fuzzy subsets) over some crisp symbols. In order to deal with arbitrary words that may be not in the input alphabet of an FACW as inputs, we established the so-called retractions and generalized extensions of FACWs by exploiting the methodology of fuzzy control.

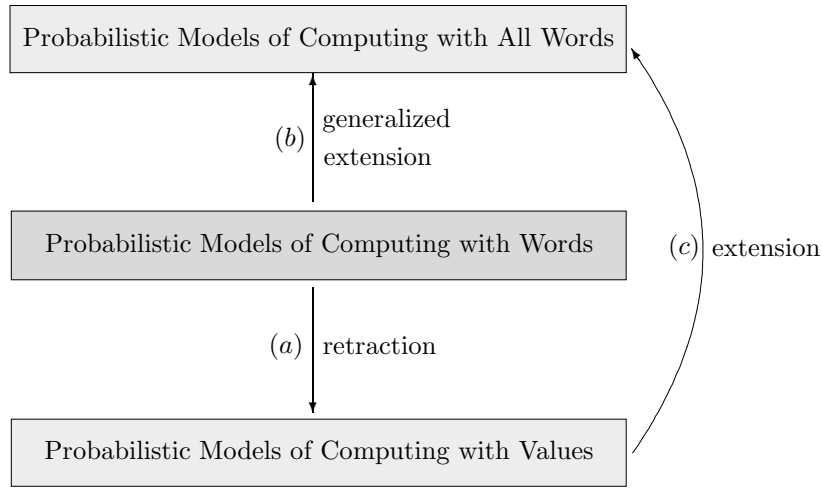

Figure 1: Interrelation among retractions, extensions, and generalized extensions.

As mentioned above, the probabilistic models of computing with words in [12] also suffer from the dependence on the underlying classical computational models. The purpose of this paper is thus to build a general probabilistic model of computing with words. We introduce probabilistic automata for computing with words (PACWs) and as well probabilistic grammars for computing with words (PGCWs) to model formally computing with words in a probabilistic framework. Probabilistic automata and probabilistic grammars have been studied since the early 1960s [11]. Relevant to our line of interest is the work of Rabin [13]. In the present paper, the words that represent generalized constraints are interpreted as probabilistic distributions or possibility distributions over some crisp symbols.

We may think that PACWs and PGCWs are specified by experts, in which only finite words are considered. For example, an expert may express his opinion on a repeated risk investment of a firm in the following proposition: If the firm is in a good situation and if it invests in the projects $A$ and $B$ with probabilities 0.7 and 0.3, respectively, then it will be still in the good situation with probability 0.9 while in a bad situation with probability 0.1. Based upon some analogous propositions, we can build a PACW or PGCW to represent the expert's opinions. In practice, in many areas expert's opinions may be naturally expressed in terms of linguistic uncertainties. Clearly, it is desirable if from such knowledge, we can make inferences about some particular actions that are not specified by the experts. For instance, one may want to assess the situations of the firm when it invests in the projects $A$ and $B$ with probabilities 0.75 and 0.25, respectively, or it invests in the project $A$ with probability 1. This motivates us to consider the so-called retractions and generalized extensions of PACWs and PGCWs.

Roughly speaking, the retraction of a PACW is a probabilistic automaton, called probabilistic automaton for computing with values (PACV), that has crisp symbols as inputs; while the generalized extension of a PACW is another probabilistic automaton, called probabilistic automaton for computing with all words (PACAW), that can accept any words as inputs (see Figure 1). As we will see, the extension from PACVs to PACAWs developed in [12] is a special case of generalized extensions. By probabilistic conditioning rather than the methodology of fuzzy control used in [2], we establish a retraction principle from computing with words to computing with values for dealing with crisp inputs and a generalized extension principle from computing with words to computing with all words for dealing with arbitrary inputs. These principles show that in the probabilistic framework, computing with values and computing with all words can be respectively implemented by computing with some special words. From a modeling viewpoint, the generalized extensions enable infinitely possible inputs to be represented by finite inputs by means of interpolation. Analogously, we investigate the retractions and the generalized extensions of PGCWs. Furthermore, we show that the retractions and the generalized extensions preserve all the three kinds of equivalences among PACWs and PGCWs consisting of the equivalence between PACWs, the equivalence between PGCWs, and the equivalence between PACWs and PGCWs.

The present work is developed closely along the lines of [2], because in our opinion, like the studies on probabilistic automata and fuzzy automata in history, the probabilistic models of computing with words deserve a study similar to that of fuzzy model of computing with words. It is worth noting that although probabilistic automata and fuzzy automata are formally similar, they have different semantics and can satisfy diverse applications; see, for example, [8, 10, 11] and the bibliographies therein. Based on the complementarity of fuzzy logic and probability theory (see [18, 24]), probabilistic models and fuzzy models for computing with words may complement each other. In the paper, we pay more attention to some aspects that are not or may not be considered for fuzzy models in [2], such as formal grammars, the linearity of the transition probability function of a generalized extension, and analytical properties comparing the transition probabilities of two near inputs.

The rest of the paper is organized as follows. In Section 2, after presenting some preliminaries in probabilistic automata and probabilistic grammars, we introduce two probabilistic models of computing with words, PACWs and PGCWs. The retractions of PACWs are established in Section 3. We develop the generalized extensions of PACWs and discuss some related analytical properties in Section 4. Section 5 is concerned with the retractions and the generalized extensions of PGCWs, and Section 6 is devoted to the equivalence preservation under the retractions and the generalized extensions. Some relationships among the retractions, the generalized extensions, and the extensions in [12] are explored in Section 7. We conclude the paper and identify some future research directions in Section 8.

## 2   Probabilistic models of computing with words

After recalling some basics of probabilistic automata and the extensions in [12], we introduce the notion of probabilistic automata for computing with words in Section 2.1. In a parallel way, we define probabilistic grammars for computing with words in Section 2.2. For later need, the equivalence between probabilistic automata and probabilistic grammars is briefly reviewed in Section 2.3.

## 2.1 Probabilistic automata for computing with words

To introduce a formal probabilistic model of computing with words, let us first review some notions on probabilistic automata.

We begin with some notations. Let $\Omega$ be a finite set. A function $\mu$ from $\Omega$ to the closed unit interval $[0,1]$ is called a *probability distribution* on $\Omega$ if $\sum_{x\in\Omega}\mu(x) = 1$. The set $\{x\in\Omega : \mu(x) > 0\}$ is called the *support* of $\mu$ and is denoted by $\mathrm{supp}(\mu)$. For any $x\in\Omega$, we use $\hat{x}$ to denote the unique probability distribution with $\hat{x}(x) = 1$, also known as the *Dirac distribution* for $x$. By $\mathcal{D}(\Omega)$ we denote the set of all probability distributions on the set $\Omega$. If $\mu$ is a probability distribution with support $\{x_1, \ldots, x_n\}$, we sometimes write $\mu$ in Zadeh's notation [25] as

$$\mu = \mu(x_1)\backslash x_1 + \mu(x_2)\backslash x_2 + \cdots + \mu(x_n)\backslash x_n.$$

With this notation, $\hat{x} = 1\backslash x$. For any $\lambda \in [0,1]$ and $\mu \in \mathcal{D}(\Omega)$, we define a *scalar multiplication* $\lambda \cdot \mu : \Omega \longrightarrow [0,1]$ by $(\lambda \cdot \mu)(x) = \lambda \cdot \mu(x)$ for all $x\in\Omega$, where the dot "$\cdot$" in $\lambda \cdot \mu(x)$ stands for the product of $\lambda$ and $\mu(x)$. We abuse the notation "$\cdot$" from now on, since the context will avoid ambiguity. Clearly, the scalar multiplication $\lambda \cdot \mu$ is not necessarily a probability distribution on $\Omega$.

For later need, let us briefly review the notion of fuzzy subsets. Each *fuzzy subset* (or simply *fuzzy set*), $\mathcal{A}$, is defined in terms of a relevant universal set $X$ by a function assigning to each element $x$ of $X$ a value $\mathcal{A}(x)$ in $[0,1]$. A fuzzy subset of $X$ can be used to formally represent a possibility distribution on $X$. We denote by $\mathcal{F}(X)$ the set of all fuzzy subsets of $X$.

Recall that a *deterministic finite automaton* is a five-tuple $A = (Q, \Sigma, \delta, q_0, F)$, where $Q$ is a finite set of states, $\Sigma$ is a finite input alphabet, $q_0 \in Q$ is the initial state, $F \subseteq Q$ is the set of final states, and $\delta$ is a mapping from $Q\times\Sigma$ to $Q$. The language accepted by $A$ is defined as $L(A) = \{s \in \Sigma^* : \delta(q_0, s) \in F\}$. As a generalization of deterministic finite automata, Rabin introduced the following probabilistic automata in the early 1960s [13].

**Definition 2.1.** A *probabilistic automaton* is a five-tuple $M = (Q, \Sigma, \delta, q_0, F)$, where:

(a) $Q$ is a finite set of states;

(b) $\Sigma$ is a finite input alphabet;

(c) $q_0 \in Q$ is the initial state;

(d) $F \subseteq Q$ is the set of final states;

(e) $\delta$, the transition probability function, is a function from $Q \times \Sigma$ to $\mathcal{D}(Q)$ that takes a state in $Q$ and an input symbol in $\Sigma$ as arguments and returns a probability distribution on $Q$.

When the probabilistic automaton is in state $p \in Q$ and if the input is $\sigma \in \Sigma$, then it can go into any one of the states $q \in Q$, and the probability of going into $q$ is $\delta(p, \sigma)(q)$. Thus, the probabilistic automaton has a define transition probability for entering state $q$ from state $p$ when receiving a string (i.e., a sequence of inputs). To give this probability, we define inductively an *extended transition probability function* from $Q \times \Sigma^*$ to $\mathcal{D}(Q)$, denoted by the same notation $\delta$, as follows:

$$\begin{aligned}\delta(p, \epsilon) &= 1\backslash p \\ \delta(p, s\sigma) &= \sum_{q\in Q} \delta(p, s)(q) \cdot \delta(q, \sigma)\end{aligned}$$

for all $s \in \Sigma^*$ and $\sigma \in \Sigma$, where $\Sigma^*$ consists of all finite strings (including the empty string $\epsilon$) over $\Sigma$, $1\backslash p$ is the Dirac distribution for $p$, and $\delta(p,s)(q) \cdot \delta(q,\sigma)$ is the scalar multiplication of the scalar $\delta(p,s)(q)$ and the probability distribution $\delta(q,\sigma)$. It is not hard to check that $\delta(p, s\sigma)$ is a probability distribution on $Q$.

The *language* accepted by the probabilistic automaton $M$ is defined as a function $L(M) : \Sigma^* \longrightarrow [0,1]$ in the following way: For any $s \in \Sigma^*$,

$$L(M)(s) = \sum_{q \in F} \delta(q_0, s)(q).$$

The value $L(M)(s)$ is exactly the probability for $M$, when started in $q_0$, to go into a state in $F$ by the input $s$, called the *accepting probability* of $s$ by $M$.

The symbols from an input alphabet are usually viewed as exact input values. In this sense, the above definition provides a model of computing with values through probabilistic automata. Therefore, we shall refer to the probabilistic automaton in Definition 2.1 as a *probabilistic automaton for computing with values* (or PACV for short). In contrast, words in the natural languages are the descriptions of some imprecise values; they can be formally represented as probability distributions or possibility distributions over a certain underlying set. Following Zadeh's opinion in [25], we interpret a word over an input alphabet $\Sigma$ as a probability distribution (resp. a possibility distribution) on $\Sigma$. In this sense, computing with words in this paper is concerned with formal computation whose input is a string of probability distributions (resp. possibility distributions) on an underlying input alphabet, instead of a string of symbols from the underlying input alphabet.

In the literature of formal language theory, a string is often called a "word". To avoid confusion in the present paper, we do not use the term "word" in this way and only use it to refer to what we mean by "word" in the phrase "computing with words." For clarity, we develop our work along the probability interpretation of words and point out several necessary modifications required to deal with the possibility interpretation of words. Thus, in what follows the term "word" means a probability distribution, unless otherwise specified.

Motivated by Ying's formal model of fuzzy automata for computing with words [22], Qiu and Wang [12] proposed a probabilistic model of computing with words via probabilistic automata. This was done by extending further the transition probability function of a PACV as follows.

**Definition 2.2.** Let $M = (Q, \Sigma, \delta, q_0, F)$ be a PACV.

(a) To handle words as inputs, $\delta$ is extended to a function from $Q \times \mathcal{D}(\Sigma)$ to $\mathcal{D}(Q)$, denoted $\hat{\delta}$, as follows:

$$\hat{\delta}(p, W) = \sum_{\sigma \in \Sigma} W(\sigma) \cdot \delta(p, \sigma)$$

for any $p \in Q$ and $W \in \mathcal{D}(\Sigma)$. It is easy to verify that $\hat{\delta}(p, W) \in \mathcal{D}(Q)$, and we thus get a probabilistic automaton $\hat{M} = (Q, \mathcal{D}(\Sigma), \hat{\delta}, q_0, F)$ with an infinite input alphabet. With the interpretation of words in terms of probability distributions, $\hat{M}$ can serve as a probabilistic, formal model of computing with all words over $\Sigma$. For convenience, we sometimes refer to $\hat{M}$ as the *extension* of $M$, which corresponds to the process $(c)$ in Figure 1.

(b) To handle strings of words as inputs, $\hat{\delta}$ in (a) is further extended to a function from $Q \times \mathcal{D}(\Sigma)^*$ to $\mathcal{D}(Q)$, denoted again by $\hat{\delta}$, as follows:

$$\hat{\delta}(p, \epsilon) = 1\backslash p$$
$$\hat{\delta}(p, SW) = \sum_{q \in Q} \hat{\delta}(p, S)(q) \cdot \hat{\delta}(q, W)$$

for all $S \in \mathcal{D}(\Sigma)^*$ and $W \in \mathcal{D}(\Sigma)$.

(c) The *word language* $L_w(\hat{M})$ accepted by $\hat{M}$ is a function from $\mathcal{D}(\Sigma)^*$ to $[0,1]$ defined by

$$L_w(\hat{M})(S) = \sum_{q \in F} \hat{\delta}(q_0, S)(q)$$

for all $S \in \mathcal{D}(\Sigma)^*$. The numerical value $L_w(\hat{M})(S)$ is the probability that $\hat{M}$ accepts the string of words $S$.

As we see from Definition 2.2, the probabilistic model of computing with words is essentially a probabilistic automaton which is the same as the probabilistic automaton in Definition 2.1. Importantly, however, the strings of inputs are different: In Definition 2.1 they are strings of values, whereas in Definition 2.2 they are strings of words. It is worth noting that the input alphabet of $\hat{M}$ consists of all probability distributions on $\Sigma$ and the transition probability function $\hat{\delta}$ in (a) of Definition 2.2 depends on the underlying probabilistic automaton $M$ as well. For this reason, we introduce a somewhat general probabilistic model of computing with words.

**Definition 2.3.** A *probabilistic automaton for computing with words* (or PACW for short) is a probabilistic automaton $M_w = (Q, \Sigma_w, \delta, q_0, F)$, where the components $Q, q_0, F$ have their same interpretation as in Definition 2.1 and the following hold:

($b'$) $\Sigma_w$ is a finite subset of $\mathcal{D}(\Sigma)$, where $\Sigma$ is a finite set of symbols, called the underlying input alphabet.

($e'$) $\delta$ is a transition probability function from $Q \times \Sigma_w$ to $\mathcal{D}(Q)$.

The new features of the model in Definition 2.3 are that the input alphabet consists of some (not necessarily all) probability distributions over a finite set of symbols (i.e., the underlying input alphabet) and the transition probability function can be specified arbitrarily. In particular, when $\Sigma_w = \mathcal{D}(\Sigma)$, we say that the PACW is a *probabilistic automaton for computing with all words* (or PACAW for short). The choice of $\Sigma_w$ and the specification of the transition probability function $\delta$ are provided by expert from experiment or intuition. The definition of language accepted by a PACV is applicable to PACWs, and we thus get a direct way of computing the string of words.

The following is a simple example coming from game theory. The reader who is not familiar with basic notions of game theory is referred to the standard textbook [5].

**Example 2.4.** Let us see the famous *prisoner's dilemma* game. It goes as follows: Two suspects have been accused of collaborating in a crime. They are in separate jail cells and cannot communicate with each other. Each has been asked to confess. If both suspects confess, each will receive a prison term of 3 years. If neither confesses, both will be released on account of insufficient evidence. On the other hand, if one suspect confesses and the other does not, the one who confesses will receive a term of only 1 year, while the other will go to prison for 5 years. The payoff matrix in Table 1 summarizes the possible outcomes, where, for example, the entry in the upper left-hand corner means a three-year sentence for each suspect.

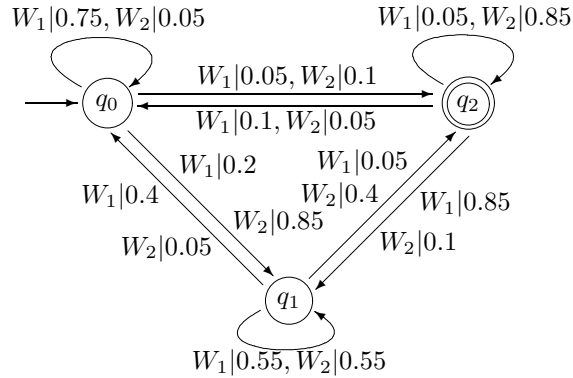

Figure 2: A probabilistic automaton for computing with words.

|  | Confess | Do not confess |
|---|---|---|
| Confess | $-3, -3$ | $-1, -5$ |
| Do not confess | $-5, -1$ | $0, 0$ |

Table 1: Payoff matrix for prisoner's dilemma.

As the table shows, every suspect faces a dilemma. If they could both agree not to confess, then each would be released. But they cannot talk to each other, and even if they could, they may not trust each other. If one of them does not confess, he risks being taken advantage of by his former accomplice. In fact, the prisoner's dilemma game is a model of many situations in real life. For example, in oligopolistic markets, firms often find themselves in a prisoner's dilemma game when making output or pricing decisions. We suppose that the two suspects may be accused many times (i.e., they are playing *repeated games*); for simplicity, we also assume that they prefer to play with *tit-for-tat* strategy: each suspect starts by "Do not confess," and thereafter prefers to choose in round $j+1$ the action chosen by his accomplice in round $j$.

We want to give a PACW to describe a suspect's dilemma. By assumption, the suspect being described can merely consider three mental states, say $q_0$, $q_1$, and $q_2$. For instance, $q_0$ might say "neither will confess", $q_1$ might say "only one will confess", and $q_2$ might say "both will confess." We use $a$ and $b$ to denote the strategies "Do not confess" and "Confess", respectively. Because the suspect is dilemmatic, it is difficult for him to make a specific choice among $a$ and $b$. We thus suppose that the suspect makes a random choice among the two possible strategies, based on a set of chosen probabilities. In other words, the suspect is supposed to adopt *mixed strategies*. (Strategies of this kind arise naturally in repeated games.) For instance, we consider two mixed strategies $W_1 = 0.9\backslash a + 0.1\backslash b$ and $W_2 = 0.1\backslash a + 0.9\backslash b$. The strategy $W_1$ means that the suspect choices $a$ with probability 0.9, while $W_2$ means that the suspect choices $b$ with probability 0.9. The transition probabilities in Figure 2 describe the suspect's belief change with his strategies. For example, the directed cycle with label $W_1|0.75$ means that the suspect believes his state is $q_0$ with probability 0.75 if he choices the mixed strategy $W_1$ at the initial state $q_0$, and actually, the game may not start.

Let $\Sigma = \{a, b\}$. Then $W_1$ and $W_2$ are two words over $\Sigma$. Take $Q = \{q_0, q_1, q_2\}$, $\Sigma_w = \{W_1, W_2\}$, and $F = \{q_2\}$. The transition probability function $\delta : Q \times \Sigma_w \longrightarrow \mathcal{D}(Q)$ follows from Figure 2. We then get a PACW $(Q, \Sigma_w, \delta, q_0, F)$, denoted by $\mathcal{M}$. The word language accepted by $\mathcal{M}$ is given by $L_w(\mathcal{M})(S) = \delta(q_0, S)(q_2)$ for all $S \in \Sigma_w^*$. For example,

$$L_w(\mathcal{M})(W_1 W_1) = 0.05, \qquad L_w(\mathcal{M})(W_1 W_2) = 0.1975,$$
$$L_w(\mathcal{M})(W_2 W_1) = 0.05, \qquad L_w(\mathcal{M})(W_2 W_2) = 0.43,$$

where $L_w(\mathcal{M})(W_2 W_2) = 0.43$ means that the suspect believes his state is being $q_2$ with probability 0.43 when he chose the mixed strategy $W_2$ in the first accusation and choices the same one in the second accusation.

We end this subsection by extending Definition 2.3 to the case of possibility distributions.

**Remark 2.5.** If the words in Definition 2.3 are interpreted as possibility distributions over $\Sigma$ (i.e., fuzzy subsets of $\Sigma$), then after replacing $\mathcal{D}(\Sigma)$ with $\mathcal{F}(\Sigma)$, the definitions of PACWs and PACAWs are still appropriate. In terms of mathematical expressions, one can regard every probability distribution as a special possibility distribution, but their semantics are clearly different.

## 2.2 Probabilistic grammars for computing with words

Before introducing probabilistic grammars for computing with words, let us recall several definitions (see, for example, [6]).

**Definition 2.6.** A *grammar* is a tuple $G = (V, \Sigma, P, S)$ where $V$ and $\Sigma$ are respectively finite sets of variables and terminals with $V \cap \Sigma = \emptyset$, $P$ is the set of productions of the form $\alpha \to \beta$, and $S \in V$ is the starting variable.

The following are some frequently used notations on grammar $G$:

1) $\Sigma^*$ is the set of all finite-length strings of $\Sigma$ (including the empty string $\epsilon$), and $\Sigma^+ = \Sigma^* \setminus \{\epsilon\}$.

2) $\eta \xrightarrow[G]{} \gamma$ means that there exist $\omega_1, \omega_2 \in (V \cup \Sigma)^*$ and $\alpha \to \beta \in P$ such that $\eta = \omega_1 \alpha \omega_2$ and $\gamma = \omega_1 \beta \omega_2$.

3) $\eta \xrightarrow[G]{*} \gamma$ denotes that there is a sequence of strings $\xi_1, \ldots, \xi_n$ such that $\xi_1 = \eta$, $\xi_n = \gamma$, and $\xi_i \xrightarrow[G]{} \xi_{i+1}$ for all $1 \leq i \leq n - 1$.

4) The language generated by the grammar $G$ is defined as $L(G) = \{s \in \Sigma^* : S \xrightarrow[G]{*} s\}$.

The name $G$ below the arrows will be omitted if the grammar $G$ that is being used is obvious. The form of productions determines the type of a grammar. It is well known that regular grammars are equivalent to deterministic finite automata, context-free grammars are equivalent to pushdown automata, and context-sensitive grammars are equivalent to Turing machines.

There are a large number of probabilistic versions arising from these grammars (see [4, 3, 14, 20], and others). For our purpose of illustrating the idea of computing with words via grammars, we focus on the following probabilistic grammar, which is closely relevant to probabilistic automata being considered in the paper.

**Definition 2.7.** A *probabilistic grammar* is a grammar $G = (V, \Sigma, P, S)$, where each production is endowed with a probability subject to the following:

- for any $A$ and $a \in \Sigma$, $\sum_B Pr(A \to aB) = 1$, where the sum runs over $\{B \in V : A \to aB \in P\}$;
- $Pr(A \to \epsilon) \in \{0, 1\}$;
- $Pr(A \to B) = 0$.

In the light of the first condition above, we see that $Pr(A \to aB) = Pr(B|A, a)$, so we sometimes adopt the latter notation. Unlike usual grammars, a probabilistic grammar accepts every string with a probability.

**Definition 2.8.** Let $G = (V, \Sigma, P, S)$ be a probabilistic grammar. The language $L(G)$ generated by $G$ is a function from $\Sigma^*$ to $[0, 1]$ defined by

$$L(G)(\epsilon) = Pr(S \to \epsilon);$$

$$L(G)(s) = \sum_{\substack{A_1, \ldots, A_l \in V \\ A_l \to \epsilon \in P}} \prod_{i=1}^{l} Pr(A_i | A_{i-1}, a_i)$$

for any string $s = a_1 \cdots a_l \in \Sigma^+$, in which $A_0 = S$.

Similar to the definition of PACWs, we have the following notion.

**Definition 2.9.** A *probabilistic grammar for computing with words* (PGCW) is a probabilistic grammar $G_w = (V, \Sigma_w, P, S)$, where all components have their same interpretation as in Definition 2.7, except that $\Sigma_w$ is now a finite set of probabilistic distributions over some underlying terminal set $\Sigma$.

The language generated by a PGCW $G_w$, called *word language* and denoted $L_w(G_w)$, is defined in the same way as in Definition 2.8.

The following is an example of PGCW arising from Example 2.4.

**Example 2.10.** Let $\Sigma = \{a, b\}$, $W_1 = 0.9\backslash a + 0.1\backslash b$, and $W_2 = 0.1\backslash a + 0.9\backslash b$. Set $V = \{q_0, q_1, q_2\}$, $\Sigma_w = \{W_1, W_2\}$, $S = \{q_0\}$, and

$$P = \{q_i \to W_k q_j : i, j \in \{0, 1, 2\}, k \in \{1, 2\}\} \cup \{q_i \to \epsilon : i = 0, 1, 2\}$$

with the probabilities

$$\begin{aligned}
Pr(q_0|q_0, W_1) &= 0.75, & Pr(q_1|q_0, W_1) &= 0.2, & Pr(q_2|q_0, W_1) &= 0.05, \\
Pr(q_0|q_1, W_1) &= 0.4, & Pr(q_1|q_1, W_1) &= 0.55, & Pr(q_2|q_1, W_1) &= 0.05, \\
Pr(q_0|q_2, W_1) &= 0.1, & Pr(q_1|q_2, W_1) &= 0.85, & Pr(q_2|q_2, W_1) &= 0.05, \\
Pr(q_0|q_0, W_2) &= 0.05, & Pr(q_1|q_0, W_2) &= 0.85, & Pr(q_2|q_0, W_2) &= 0.1, \\
Pr(q_0|q_1, W_2) &= 0.05, & Pr(q_1|q_1, W_2) &= 0.55, & Pr(q_2|q_1, W_2) &= 0.4, \\
Pr(q_0|q_2, W_2) &= 0.05, & Pr(q_1|q_2, W_2) &= 0.1, & Pr(q_2|q_2, W_2) &= 0.85, \\
Pr(q_0 \to \epsilon) &= 0, & Pr(q_1 \to \epsilon) &= 0, & Pr(q_2 \to \epsilon) &= 1.
\end{aligned}$$

We then get a PGCW $(V, \Sigma_w, P, S)$, denoted by $\mathcal{G}$. It is easy to verify that $L_w(\mathcal{G})(s) = L_w(\mathcal{M})(s)$ for all $s \in \Sigma_w^*$, where $\mathcal{M}$ is the PACW in Example 2.4.

## 2.3 Probabilistic automata vs. probabilistic grammars

It is not hard to check that probabilistic automata in Definition 2.1 (i.e., PACVs) and probabilistic grammars in Definition 2.7 are equivalent. For later need, we record a construction of the equivalence.

Given a probabilistic grammar $G = (V, \Sigma, P, S)$, the following process generates a probabilistic automaton $M_G = (Q, \Sigma, \delta, q_0, F)$ satisfying that $L(G)(s) = L(M_G)(s)$ for all $s \in \Sigma^*$:

1) Let $Q = V$, $q_0 = S$, and $F = \{A \in V : Pr(A \to \epsilon) = 1\}$.

2) Define $\delta(A, a)(B) = Pr(B|A, a)$ for all $A, B \in Q$ and $a \in \Sigma$.

In turn, given a probabilistic automaton $M = (Q, \Sigma, \delta, q_0, F)$, we can also construct an equivalent probabilistic grammar $G_M = (V, \Sigma, P, S)$:

1) Let $V = Q$ and $S = q_0$.

2) Let $P = \{A \to aB : A, B \in V, a \in \Sigma\} \cup \{A \to \epsilon : A \in V\}$ and define the probabilities of the productions as follows:

$$Pr(A \to aB) = \delta(A, a)(B);$$

$$Pr(A \to \epsilon) = \begin{cases} 1, & \text{if } A \in F \\ 0, & \text{otherwise.} \end{cases}$$

For convenience, we say that $M_G$ is the probabilistic automaton induced from the probabilistic grammar $G$ and also $G_M$ is the probabilistic grammar induced from the probabilistic automaton $M$. Clearly, the construction above is applicable to PACWs and PGCWs, which gives the equivalence between them.

## 3 Retractions of PACWs: Towards computing with values

Recall that the probabilistic model of computing with words derived by Qiu and Wang is in fact an extension from computing with values to computing with all words. In this section, we in turn address how to tackle computing with values when we only have a probabilistic automaton $M_w = (Q, \Sigma_w, \delta, q_0, F)$ for computing with words. To this end, we shall establish a probabilistic automaton $M_w^\downarrow = (Q, \Sigma, \delta^\downarrow, q_0, F)$, where the components $Q, q_0, F$ are the same as those of $M_w$, $\Sigma$ is the underlying input alphabet of $M_w$, and $\delta^\downarrow$ which depends on the transition probability function of $M_w$ need to be defined.

For any $p, q \in Q$ and $\sigma \in \Sigma$, we want to derive a formula for computing $\delta^\downarrow(p, \sigma)(q)$ by conditional probability. Let $C, N$, and $I$ denote the random variables of current state, the next state, and real input (crisp input), respectively. Let $O$ represent what we observe about the current input. Given a PACW $M_w = (Q, \Sigma_w, \delta, q_0, F)$, we can extract the following information about conditional probability: $Pr(N = q|O = W, C = p) = \delta(p, W)(q)$ and $Pr(I = \sigma|O = W) = W(\sigma)$. Further, we make several natural assumptions:

(a) The prior probabilities $Pr(W)$ are equal for all $W \in \Sigma_w$.

(b) Given the current state and the observation, the next state is independent with the real input.

(c) Given the observation, the real input is independent with the current state.

With these assumptions, we have the following calculation.

$$\begin{aligned}
\delta^{\downarrow}(p,\sigma)(q) &= Pr(N=q|I=\sigma, C=p) \\
&= \frac{Pr(N=q, I=\sigma|C=p)}{Pr(I=\sigma|C=p)} \\
&= \frac{\sum_{W \in \Sigma_w} Pr(N=q, I=\sigma|O=W, C=p)Pr(O=W|C=p)}{\sum_{U \in \Sigma_w} Pr(I=\sigma|O=U, C=p)Pr(O=U|C=p)} \\
&= \frac{\sum_{W \in \Sigma_w} Pr(N=q, I=\sigma|O=W, C=p)}{\sum_{U \in \Sigma_w} Pr(I=\sigma|O=U, C=p)} \quad \text{(by assumption (a))} \\
&= \frac{\sum_{W \in \Sigma_w} Pr(N=q|O=W, C=p)Pr(I=\sigma|O=W, C=p)}{\sum_{U \in \Sigma_w} Pr(I=\sigma|O=U, C=p)} \\
&\qquad\qquad\qquad\qquad\qquad\qquad\qquad\qquad \text{(by assumption (b))} \\
&= \frac{\sum_{W \in \Sigma_w} Pr(N=q|O=W, C=p)Pr(I=\sigma|O=W)}{\sum_{U \in \Sigma_w} Pr(I=\sigma|O=U)} \quad \text{(by assumption (c))} \\
&= \frac{\sum_{W \in \Sigma_w} Pr(N=q|O=W, C=p)W(\sigma)}{\sum_{U \in \Sigma_w} U(\sigma)} \\
&= \sum_{W \in \Sigma_w} \frac{W(\sigma)}{\sum_{U \in \Sigma_w} U(\sigma)} Pr(N=q|O=W, C=p) \\
&= \sum_{W \in \Sigma_w} \frac{W(\sigma)}{\sum_{U \in \Sigma_w} U(\sigma)} \delta(p, W)(q),
\end{aligned}$$

namely,

$$\delta^{\downarrow}(p,\sigma)(q) = \sum_{W \in \Sigma_w} \frac{W(\sigma)}{\sum_{U \in \Sigma_w} U(\sigma)} \delta(p, W)(q).$$

Since the definition of $\delta^{\downarrow}$ follows from a conditional probability, it is clear that for any $p \in Q$ and $\sigma \in \Sigma$, $\delta^{\downarrow}(p, \sigma)$ is a probability distribution on $Q$. Based upon the transition probability function $\delta^{\downarrow}$, we can get a PACV from $M_w$ as follows.

**Definition 3.1.** Let $M_w = (Q, \Sigma_w, \delta, q_0, F)$ be a PACW. The *retraction* of $M_w$ is a PACV $M_w^{\downarrow} = (Q, \Sigma, \delta^{\downarrow}, q_0, F)$, where:

($a$) The components $Q, q_0, F$ are the same as those of $M_w$.

(b) $\Sigma$ is the underlying input alphabet of $M_w$.

(c) $\delta^\downarrow$, the transition probability function, is a mapping from $Q \times \Sigma$ to $\mathcal{D}(Q)$ that maps $(p,\sigma) \in Q \times \Sigma$ to a probability distribution $\delta^\downarrow(p,\sigma)$ on $Q$ defined by

$$\delta^\downarrow(p,\sigma)(q) = \sum_{W \in \Sigma_w} \frac{W(\sigma)}{\sum_{U \in \Sigma_w} U(\sigma)} \cdot \delta(p,W)(q) \tag{1}$$

for any $q \in Q$.

Notice that for a given $M_w$, the coefficient $W(\sigma)/\sum_{U \in \Sigma_w} U(\sigma)$ in the above equation (1) is only dependent on $W$ and $\sigma$, so for convenience, we will always write $\chi_\sigma(W)$ for $W(\sigma)/\sum_{U \in \Sigma_w} U(\sigma)$ in the rest of this paper. With this notation, (1) in Definition 3.1 is as follows:

$$\delta^\downarrow(p,\sigma)(q) = \sum_{W \in \Sigma_w} \chi_\sigma(W) \cdot \delta(p,W)(q). \tag{1'}$$

The retraction of $M_w$ deals with exact inputs, and thus it may serve as a device for computing with values. We will refer to "$\downarrow$" as the operation of obtaining the retraction. As an example, we now derive the retraction of the PACW $\mathcal{M}$ given in Example 2.4.

**Example 3.2.** Consider the PACW $\mathcal{M}$ in Example 2.4. By definition, we see that

$$\chi_a(W_1) = \frac{W_1(a)}{W_1(a)+W_2(a)} = 0.9, \qquad \chi_a(W_2) = \frac{W_2(a)}{W_1(a)+W_2(a)} = 0.1,$$

$$\chi_b(W_1) = \frac{W_1(b)}{W_1(b)+W_2(b)} = 0.1, \qquad \chi_b(W_2) = \frac{W_2(b)}{W_1(b)+W_2(b)} = 0.9.$$

Further, using (1') yields that

$$\begin{aligned}
\delta^\downarrow(q_0,a) &= \chi_a(W_1) \cdot \delta(q_0,W_1) + \chi_a(W_2) \cdot \delta(q_0,W_2) = 0.68\backslash q_0 + 0.265\backslash q_1 + 0.055\backslash q_2, \\
\delta^\downarrow(q_0,b) &= \chi_b(W_1) \cdot \delta(q_0,W_1) + \chi_b(W_2) \cdot \delta(q_0,W_2) = 0.12\backslash q_0 + 0.785\backslash q_1 + 0.095\backslash q_2, \\
\delta^\downarrow(q_1,a) &= \chi_a(W_1) \cdot \delta(q_1,W_1) + \chi_a(W_2) \cdot \delta(q_1,W_2) = 0.365\backslash q_0 + 0.55\backslash q_1 + 0.085\backslash q_2, \\
\delta^\downarrow(q_1,b) &= \chi_b(W_1) \cdot \delta(q_1,W_1) + \chi_b(W_2) \cdot \delta(q_1,W_2) = 0.085\backslash q_0 + 0.55\backslash q_1 + 0.365\backslash q_2, \\
\delta^\downarrow(q_2,a) &= \chi_a(W_1) \cdot \delta(q_2,W_1) + \chi_a(W_2) \cdot \delta(q_2,W_2) = 0.095\backslash q_0 + 0.775\backslash q_1 + 0.13\backslash q_2, \\
\delta^\downarrow(q_2,b) &= \chi_b(W_1) \cdot \delta(q_2,W_1) + \chi_b(W_2) \cdot \delta(q_2,W_2) = 0.055\backslash q_0 + 0.175\backslash q_1 + 0.77\backslash q_2.
\end{aligned}$$

This transition probability function $\delta^\downarrow$, together with some data of $\mathcal{M}$, gives rise to $\mathcal{M}^\downarrow = (\{q_0,q_1,q_2\},\{a,b\},\delta^\downarrow,q_0,\{q_2\})$. The language accepted by $\mathcal{M}^\downarrow$ is defined by $L(\mathcal{M}^\downarrow)(s) = \delta^\downarrow(q_0,s)(q_2)$ for all $s \in \{a,b\}^*$. For example, $L(\mathcal{M}^\downarrow)(ab) = \delta(q_0,ab)(q_2) = 0.203675$.

We end this subsection by making a close link between computing with values and computing with words.

**Theorem 3.3.** Suppose that $M_w = (Q,\Sigma_w,\delta,q_0,F)$ is a PACW and $M_w^\downarrow = (Q,\Sigma,\delta^\downarrow,q_0,F)$ is the retraction of $M_w$. Then for any $s = \sigma_1 \cdots \sigma_l \in \Sigma^*$, we have that

$$L(M_w^\downarrow)(s) = \sum_{W_1,\ldots,W_l \in \Sigma_w} L_w(M_w)(W_1 \cdots W_l) \cdot \prod_{i=1}^l \chi_{\sigma_i}(W_i),$$

where $\chi_{\sigma_i}(W_i) = W_i(\sigma_i)/\sum_{U \in \Sigma_w} U(\sigma_i)$.

The above theorem may be seen as a retraction principle from computing with words to computing with values. The meaning of this theorem is that computing with values can be implemented by computing with words. The advantage of this approach is that we can directly obtain the accepting probability of a string of values from the accepting probabilities of some strings of words, that is, we need not compute the transition probability function $\delta^{\downarrow}$; the price of doing so is that the number of computations for implementing computing with values by computing with words increases exponentially as the length of the input string. To illustrate this, let us revisit Example 3.2 to compute $L(\mathcal{M}^{\downarrow})(ab)$ by using the result of Theorem 3.3. By the equality in Theorem 3.3 and the calculated results in Example 2.4, we obtain that

$$
\begin{aligned}
L(\mathcal{M}^{\downarrow})(ab) &= L_w(\mathcal{M})(W_1 W_1)\chi_a(W_1)\chi_b(W_1) + L_w(\mathcal{M})(W_1 W_2)\chi_a(W_1)\chi_b(W_2) \\
&\quad + L_w(\mathcal{M})(W_2 W_1)\chi_a(W_2)\chi_b(W_1) + L_w(\mathcal{M})(W_2 W_2)\chi_a(W_2)\chi_b(W_2) \\
&= 0.05 \times 0.9 \times 0.1 + 0.1975 \times 0.9 \times 0.9 + 0.05 \times 0.1 \times 0.1 \\
&\quad + 0.43 \times 0.1 \times 0.9 \\
&= 0.203675.
\end{aligned}
$$

This coincides with the result obtaining from the transition probability function of $\mathcal{M}^{\downarrow}$.

We are now ready to prove Theorem 3.3. To this end, it is convenient to have the following lemma.

**Lemma 3.4.** *Let $M_w = (Q, \Sigma_w, \delta, q_0, F)$ be a PACW and $M_w^{\downarrow} = (Q, \Sigma, \delta^{\downarrow}, q_0, F)$ be the retraction of $M_w$. Then for any $p, q \in Q$ and $s = \sigma_1 \cdots \sigma_l \in \Sigma^*$, we have that*

$$
\delta^{\downarrow}(p, s)(q) = \sum_{W_1, \ldots, W_l \in \Sigma_w} \delta(p, W_1 \cdots W_l)(q) \cdot \prod_{i=1}^{l} \chi_{\sigma_i}(W_i).
$$

**Proof.** We prove it by induction on $l$.

1) For the basis step, namely, $l = 0$, it is trivial.

2) The induction hypothesis is that the above equation holds for $s = \sigma_1 \cdots \sigma_l$. We now prove the same for $s\sigma_{l+1}$. Using the definition of $\delta^{\downarrow}$ and the induction hypothesis, we have the following.

$$
\begin{aligned}
\delta^{\downarrow}(p, s\sigma_{l+1})(q) &= \delta^{\downarrow}(p, \sigma_1 \cdots \sigma_l \sigma_{l+1})(q) \\
&= \Big[\sum_{q' \in Q} \delta^{\downarrow}(p, \sigma_1 \cdots \sigma_l)(q') \cdot \delta^{\downarrow}(q', \sigma_{l+1})\Big](q) \\
&= \sum_{q' \in Q} \delta^{\downarrow}(p, \sigma_1 \cdots \sigma_l)(q') \cdot \delta^{\downarrow}(q', \sigma_{l+1})(q) \\
&= \sum_{q' \in Q} \Big[\sum_{W_1, \ldots, W_l \in \Sigma_w} \delta(p, W_1 \cdots W_l)(q') \cdot \prod_{i=1}^{l} \chi_{\sigma_i}(W_i)\Big] \cdot \delta^{\downarrow}(q', \sigma_{l+1})(q) \\
&= \sum_{q' \in Q} \sum_{W_1, \ldots, W_l \in \Sigma_w} \delta(p, W_1 \cdots W_l)(q') \cdot \delta^{\downarrow}(q', \sigma_{l+1})(q) \cdot \prod_{i=1}^{l} \chi_{\sigma_i}(W_i)
\end{aligned}
$$

$$
\begin{aligned}
&= \sum_{q' \in Q} \sum_{W_1, \ldots, W_l \in \Sigma_w} \delta(p, W_1 \cdots W_l)(q') \cdot \Big[ \sum_{W_{l+1} \in \Sigma_w} \chi_{\sigma_{l+1}}(W_{l+1}) \\
&\qquad\qquad\qquad\qquad\qquad\qquad\qquad \cdot \delta(q', W_{l+1})(q) \Big] \cdot \prod_{i=1}^{l} \chi_{\sigma_i}(W_i) \\
&= \sum_{q' \in Q} \sum_{W_1, \ldots, W_l \in \Sigma_w} \sum_{W_{l+1} \in \Sigma_w} \delta(p, W_1 \cdots W_l)(q') \cdot \delta(q', W_{l+1})(q) \\
&\qquad\qquad\qquad\qquad\qquad\qquad\qquad\qquad \cdot \prod_{i=1}^{l+1} \chi_{\sigma_i}(W_i) \\
&= \sum_{q' \in Q} \sum_{W_1, \ldots, W_{l+1} \in \Sigma_w} \delta(p, W_1 \cdots W_l)(q') \cdot \delta(q', W_{l+1})(q) \cdot \prod_{i=1}^{l+1} \chi_{\sigma_i}(W_i) \\
&= \sum_{W_1, \ldots, W_{l+1} \in \Sigma_w} \sum_{q' \in Q} \delta(p, W_1 \cdots W_l)(q') \cdot \delta(q', W_{l+1})(q) \cdot \prod_{i=1}^{l+1} \chi_{\sigma_i}(W_i) \\
&= \sum_{W_1, \ldots, W_{l+1} \in \Sigma_w} \Big[ \sum_{q' \in Q} \delta(p, W_1 \cdots W_l)(q') \cdot \delta(q', W_{l+1})(q) \Big] \cdot \prod_{i=1}^{l+1} \chi_{\sigma_i}(W_i) \\
&= \sum_{W_1, \ldots, W_{l+1} \in \Sigma_w} \delta(p, W_1 \cdots W_{l+1})(q) \cdot \prod_{i=1}^{l+1} \chi_{\sigma_i}(W_i),
\end{aligned}
$$

namely, $\delta^{\downarrow}(p, s\sigma_{l+1})(q) = \sum_{W_1, \ldots, W_{l+1} \in \Sigma_w} \delta(p, W_1 \cdots W_{l+1})(q) \cdot \prod_{i=1}^{l+1} \chi_{\sigma_i}(W_i)$, which proves the lemma. □

We are now in the position to verify Theorem 3.3.

**Proof of Theorem 3.3.** By the definition of $L_w(M_w)$ and Lemma 3.4, we have the equation below.

$$
\begin{aligned}
L(M_w^{\downarrow})(s) &= \sum_{q \in F} \delta^{\downarrow}(q_0, \sigma_1 \cdots \sigma_l)(q) \\
&= \sum_{q \in F} \sum_{W_1, \ldots, W_l \in \Sigma_w} \delta(q_0, W_1 \cdots W_l)(q) \cdot \prod_{i=1}^{l} \chi_{\sigma_i}(W_i) \\
&= \sum_{W_1, \ldots, W_l \in \Sigma_w} \sum_{q \in F} \delta(q_0, W_1 \cdots W_l)(q) \cdot \prod_{i=1}^{l} \chi_{\sigma_i}(W_i) \\
&= \sum_{W_1, \ldots, W_l \in \Sigma_w} \Big[ \sum_{q \in F} \delta(q_0, W_1 \cdots W_l)(q) \Big] \cdot \prod_{i=1}^{l} \chi_{\sigma_i}(W_i) \\
&= \sum_{W_1, \ldots, W_l \in \Sigma_w} L_w(M_w)(W_1 \cdots W_l) \cdot \prod_{i=1}^{l} \chi_{\sigma_i}(W_i),
\end{aligned}
$$

which completes the proof of the theorem. □

**Remark 3.5.** One can readily verify that Definition 3.1 and Theorem 3.3 remain valid without any modifications when the words are interpreted as possibility distributions.

## 4   Generalized extensions of PACWs: Towards computing with all words

Having finished the transformation from computing with words to computing with values in the preceding section, we turn our attention to another transformation which makes a PACW more robust in the sense that it can deal with more inputs. More explicitly, suppose that there is a PACW $M_w = (Q, \Sigma_w, \delta, q_0, F)$ for computing with words. Note that the input alphabet $\Sigma_w$ comprises only finite words over $\Sigma$. To allow more words as inputs, we will extend $\delta$ to a transition probability function $\delta^\uparrow : Q \times \mathcal{D}(\Sigma) \longrightarrow \mathcal{D}(Q)$. As a result, we will obtain a PACAW $M_w^\uparrow = (Q, \mathcal{D}(\Sigma), \delta^\uparrow, q_0, F)$ which can accept more words than in $\Sigma_w$ as inputs.

We derive the PACAW $M_w^\uparrow$ and discuss the computation of $L_w(M_w^\uparrow)$ in the first subsection. In Section 4.2, we examine some analytical properties of generalized extensions that compare the transition probabilities of two near inputs.

### 4.1   Generalized extensions of PACWs

Let $M_w = (Q, \Sigma_w, \delta, q_0, F)$ be a PACW. Recall that in the definition of retractions, the key ingredient is the induced transition probability function $\delta^\downarrow$. Now, we would like to use the definition of $\delta^\downarrow$ to derive the transition probability function $\delta^\uparrow$ for dealing with inputs of arbitrary words.

Let us begin with some special words, Dirac distributions. Clearly, the transition probabilities of a Dirac distribution $\hat\sigma \in \mathcal{D}(\Sigma)$ and the corresponding $\sigma \in \Sigma$ should be the same when considering them as inputs of $M_w$. In light of this, it is reasonable to define that for any $p \in Q$ and any Dirac distribution $\hat\sigma \in \mathcal{D}(\Sigma)$,

$$\delta^\uparrow(p, \hat\sigma) = \delta^\downarrow(p, \sigma).$$

We now consider the case of any word as inputs. For any $W' \in \mathcal{D}(\Sigma)$, we have that $W' = \sum_{\sigma \in \Sigma} W'(\sigma) \cdot \hat\sigma$. We thus see that the Dirac distributions $\hat\sigma$'s, $\sigma \in \Sigma$, play a role of basis. So we then extend by linearity the previous definition of $\delta^\uparrow$ for Dirac distributions to $W'$ as follows:

$$\delta^\uparrow(p, W') = \sum_{\sigma \in \Sigma} W'(\sigma) \cdot \delta^\uparrow(p, \hat\sigma). \tag{2}$$

Since $\delta^\downarrow(p, \sigma) = \sum_{W \in \Sigma_w} \chi_\sigma(W) \cdot \delta(p, W)$ by the equation (1'), we obtain that

$$\begin{aligned}
\delta^\uparrow(p, W') &= \sum_{\sigma \in \Sigma} W'(\sigma) \cdot \delta^\uparrow(p, \hat\sigma) \\
&= \sum_{\sigma \in \Sigma} W'(\sigma) \cdot \delta^\downarrow(p, \sigma) \\
&= \sum_{\sigma \in \Sigma} W'(\sigma) \cdot \Big[ \sum_{W \in \Sigma_w} \chi_\sigma(W) \cdot \delta(p, W) \Big] \\
&= \sum_{\sigma \in \Sigma} \sum_{W \in \Sigma_w} W'(\sigma) \cdot \chi_\sigma(W) \cdot \delta(p, W) \\
&= \sum_{W \in \Sigma_w} \sum_{\sigma \in \Sigma} W'(\sigma) \cdot \chi_\sigma(W) \cdot \delta(p, W) \\
&= \sum_{W \in \Sigma_w} \Big[ \sum_{\sigma \in \Sigma} W'(\sigma) \cdot \chi_\sigma(W) \Big] \cdot \delta(p, W).
\end{aligned}$$

Since $\chi_\sigma(W)$ in the above equation depends merely on $W$ and $\sigma$, it follows that the sum $\sum_{\sigma\in\Sigma} W'(\sigma) \cdot \chi_\sigma(W)$ is only dependent on $W'$ and $W$, and hence, we will always write $\theta_{W'}(W)$ for $\sum_{\sigma\in\Sigma} W'(\sigma) \cdot \chi_\sigma(W)$ for the sake of convenience. As a result, we get that
$$\delta^\uparrow(p, W') = \sum_{W\in\Sigma_w} \theta_{W'}(W) \cdot \delta(p, W).$$

Based on the definition of $\delta^\uparrow$, we have the following.

**Definition 4.1.** Let $M_w = (Q, \Sigma_w, \delta, q_0, F)$ be a PACW. The *generalized extension* of $M_w$ is a PACAW $M_w^\uparrow = (Q, \mathcal{D}(\Sigma), \delta^\uparrow, q_0, F)$, where:

(a) The components $Q, q_0, F$ are the same as those of $M_w$.

(b) $\mathcal{D}(\Sigma)$ consists of all probability distributions over the underlying input alphabet of $M_w$.

(c) $\delta^\uparrow$, the transition probability function, is a mapping from $Q \times \mathcal{D}(\Sigma)$ to $\mathcal{D}(Q)$ defined by
$$\delta^\uparrow(p, W') = \sum_{W\in\Sigma_w} \theta_{W'}(W) \cdot \delta(p, W) \qquad (3)$$
for any $(p, W') \in Q \times \mathcal{D}(\Sigma)$, where $\theta_{W'}(W) = \sum_{\sigma\in\Sigma} W'(\sigma) \cdot \left[W(\sigma)/\sum_{U\in\Sigma_w} U(\sigma)\right]$.

For any $(p, W') \in Q \times \mathcal{D}(\Sigma)$, it follows from the definition of $\delta^\uparrow$ that $\delta^\uparrow(p, W')$ is indeed a probability distribution on $Q$, so Definition 4.1 is valid. As we see from Definition 4.1, the generalized extension $M_w^\uparrow$ of $M_w$ can deal with all words over the underlying input alphabet of $M_w$ as inputs. We thus consider $M_w^\uparrow$ as a device for computing with all words and refer to "$\uparrow$" as the operation of obtaining the generalized extension.

The formula given in Definition 4.1 for computing $\delta^\uparrow(p, W')$ seems complicated. In fact, this is not a problem since one can use the equation (2) to compute it. By the argument that $\delta^\uparrow(p, \hat{\sigma}) = \delta^\downarrow(p, \sigma)$ for all $\sigma \in \Sigma$, we see that it is easy to compute $\delta^\uparrow(p, W')$ once we have obtained the retraction $M_w^\downarrow$ of $M_w$.

We now present an example to illustrate the previous definition.

**Example 4.2.** Let us first derive the generalized extension of the PACW $\mathcal{M}$ produced in Example 2.4. By Definition 4.1, we have that $\mathcal{M}^\uparrow = (\{q_0, q_1, q_2\}, \mathcal{D}(\{a, b\}), \delta^\uparrow, q_0, \{q_2\})$, where $\delta^\uparrow$ follows from the following calculation: For any $W' = \alpha\backslash a + (1-\alpha)\backslash b \in \mathcal{D}(\{a,b\})$ with $\alpha \in [0, 1]$, by the equation (2) (or equivalently, the equation (3)) we have that

$$\begin{aligned}
\delta^\uparrow(q_0, W') &= W'(a) \cdot \delta^\uparrow(q_0, \hat{a}) + W'(b) \cdot \delta^\uparrow(q_0, \hat{b}) \\
&= W'(a) \cdot \delta^\downarrow(q_0, a) + W'(b) \cdot \delta^\downarrow(q_0, b) \\
&= (0.12 + 0.56\alpha)\backslash q_0 + (0.785 - 0.52\alpha)\backslash q_1 + (0.095 - 0.04\alpha)\backslash q_2.
\end{aligned}$$

By the same token, we have the following:

$$\begin{aligned}
\delta^\uparrow(q_1, W') &= (0.085 + 0.28\alpha)\backslash q_0 + 0.55\backslash q_1 + (0.365 - 0.28\alpha)\backslash q_2, \\
\delta^\uparrow(q_2, W') &= (0.055 + 0.04\alpha)\backslash q_0 + (0.175 + 0.6\alpha)\backslash q_1 + (0.77 - 0.64\alpha)\backslash q_2.
\end{aligned}$$

The following remark justifies the name of generalized extensions.

**Remark 4.3.** We remark that the generalized extension is generally not an extension in a strictly mathematical sense, that is, there may exist $p \in Q$ and $W \in \Sigma_w$ such that $\delta^\uparrow(p, W) \neq \delta(p, W)$. For instance, in Example 4.2,

$$\delta(q_0, W_1) = 0.75\backslash q_0 + 0.2\backslash q_1 + 0.05\backslash q_2,$$

while

$$\delta^\uparrow(q_0, W_1) = 0.624\backslash q_0 + 0.317\backslash q_1 + 0.059\backslash q_2;$$

they are not equal. The appearance of such an inequality is not surprising if we have noticed that the calculation of $\delta^\uparrow(p, W)$ depends on all $W' \in \Sigma_w$ and $\delta(p, W')$, while the words in $\Sigma_w$ may be intersecting each other. Clearly, if the calculation of $\delta^\uparrow(p, W)$ is not disturbed by those $W' \in \Sigma_w \backslash \{W\}$ and $\delta(p, W')$, then the generalized extension must be an extension. For example, if each word in $\Sigma_w$ degenerates into a Dirac distribution, then it is not hard to check that the generalized extension is indeed an extension.

The equation (2) also motivates us to consider the linearity of $\delta^\uparrow(p, W')$ on the second argument. To make this precise, we need more notions. A vector is called *stochastic* if all its entries are nonnegative and the sum of its entries equals 1. Assume that $\Sigma = \{\sigma_1, \sigma_2, \ldots, \sigma_n\}$. Then each word $W$ over $\Sigma$ can be uniquely written as an $n$-dimensional stochastic row vector $[W(\sigma_1), W(\sigma_2), \ldots, W(\sigma_n)]$. A *linear combination* of some words $W_1, W_2, \ldots, W_l$ over $\Sigma$ is an expression of the form $k_1 W_1 + k_2 W_2 + \cdots + k_l W_l$, where all $k_i$'s lie in $\mathbb{R}$, the real numbers. In other words, the linear combination $k_1 W_1 + k_2 W_2 + \cdots + k_l W_l$ is the $n$-dimensional row vector $[c_1, c_2, \ldots, c_n]$ with $c_j = \sum_{i=1}^l k_i W_i(\sigma_j)$, $j = 1, \ldots, n$.

A linear combination of words does not necessarily yield a word. However, if the linear combination is indeed a word, then the transition probability when inputting the linear combination can be computed in the following ways.

**Proposition 4.4.** *Suppose that the linear combination $W' = \sum_{i=1}^l k_i W_i'$ with $W_i' \in \mathcal{D}(\Sigma)$ is a word over $\Sigma$. Then $\delta^\uparrow(p, W') = \sum_{i=1}^l k_i \cdot \delta^\uparrow(p, W_i')$.*

**Proof.** It follows from the equation (2) that

$$\begin{aligned}
\delta^\uparrow(p, W') &= \sum_{\sigma \in \Sigma} W'(\sigma) \cdot \delta^\uparrow(p, \hat{\sigma}) \\
&= \sum_{\sigma \in \Sigma} \Big[\sum_{i=1}^l k_i W_i'(\sigma)\Big] \cdot \delta^\uparrow(p, \hat{\sigma}) \\
&= \sum_{i=1}^l k_i \cdot \Big[\sum_{\sigma \in \Sigma} W_i'(\sigma) \cdot \delta^\uparrow(p, \hat{\sigma})\Big] \\
&= \sum_{i=1}^l k_i \cdot \delta^\uparrow(p, W_i'),
\end{aligned}$$

which finishes the proof of the proposition. □

The proposition above shows that $\delta^\uparrow(p, W')$ is linear on the second argument. Further, this proposition gives rise to a simple corollary which is helpful to calculate $\delta^\uparrow(p, W')$. To state this result, we appeal to a concept from linear algebra.

A set of some words over $\Sigma$ is *linearly independent* if none of its elements is a linear combination of the others. Keep the assumption that $\Sigma = \{\sigma_1, \sigma_2, \ldots, \sigma_n\}$. It then follows from the theory of linear algebra that the number of linearly independent words over $\Sigma$ is at most $n$. Further, if $W'_1, \ldots, W'_n$ are $n$ linearly independent words over $\Sigma$, then any word $W'$ over $\Sigma$ can be uniquely expressed as a linear combination of these words, that is, $W' = k_1 W'_1 + \cdots + k_n W'_n$ for some $k_i \in \mathbb{R}$. Finding $n$ linearly independent words over $\Sigma$ and expressing a word in the form of linear combination of these words are fairly routine exercises in linear algebra; we omit details here. The following corollary is a generalization of the equation (2), which allows us to compute transition probabilities from an arbitrary set of linearly independent words.

**Corollary 4.5.** *To compute $\delta^\uparrow(p, W')$ for any $(p, W') \in Q \times \mathcal{D}(\Sigma)$, we can follow the steps below:*

(1) *Find any $n$ linearly independent words over $\Sigma$, say $W'_1, \ldots, W'_n$, and write $W' = k_1 W'_1 + \cdots + k_n W'_n$.*

(2) *Compute $\delta^\uparrow(p, W'_i)$ for all $i = 1, \ldots, n$.*

(3) *Compute the sum $\sum_{i=1}^n k_i \cdot \delta^\uparrow(p, W'_i)$, which exactly equals $\delta^\uparrow(p, W')$.*

**Proof.** It follows immediately from some facts on linear algebra and Proposition 4.4. □

Analogous to Theorem 3.3, we can also establish a close link between computing with some special words (i.e., those in $\Sigma_w$) and computing with all words.

**Theorem 4.6.** *Suppose that $M_w = (Q, \Sigma_w, \delta, q_0, F)$ is a PACW and $M_w^\uparrow = (Q, \mathcal{D}(\Sigma), \delta^\uparrow, q_0, F)$ is the generalized extension of $M_w$. Then for any $S = W'_1 \cdots W'_l \in \mathcal{D}(\Sigma)^*$, we have that*

$$L_w(M_w^\uparrow)(S) = \sum_{W_1, \ldots, W_l \in \Sigma_w} L_w(M_w)(W_1 \cdots W_l) \cdot \prod_{i=1}^l \theta_{W'_i}(W_i),$$

*where $\theta_{W'_i}(W_i) = \sum_{\sigma \in \Sigma} W'_i(\sigma) \cdot \left[ W_i(\sigma) / \sum_{U \in \Sigma_w} U(\sigma) \right]$.*

Theorem 4.6 may be seen as a generalized extension principle from computing with special words to computing with all words. The meaning of this theorem is that computing with all words can be implemented by computing with special words; and thus, it gives us a way of dealing with arbitrary words as inputs of a PACW. It is clear that the number of computations for implementing computing with all words by computing with words increases exponentially as the length of the input string. To see this, let us revisit Example 4.2.

**Example 4.7.** Consider the generalized extension $\mathcal{M}^\uparrow = (\{q_0, q_1, q_2\}, \mathcal{D}(\{a, b\}), \delta^\uparrow, q_0, \{q_2\})$ in Example 4.2. As an example, taking $W' = 0.2\backslash a + 0.8\backslash b$, we now use two approaches, the definition of $\delta^\uparrow$ and Theorem 4.6, to compute $L_w(\mathcal{M}^\uparrow)(W'W')$.

By Example 4.2, we see that

$$\begin{aligned}
\delta^\uparrow(q_0, W') &= 0.232\backslash q_0 + 0.681\backslash q_1 + 0.087\backslash q_2, \\
\delta^\uparrow(q_1, W') &= 0.141\backslash q_0 + 0.55\backslash q_1 + 0.309\backslash q_2, \\
\delta^\uparrow(q_2, W') &= 0.063\backslash q_0 + 0.295\backslash q_1 + 0.642\backslash q_2.
\end{aligned}$$

Thereby, it follows from the definition of word languages that

$$\begin{aligned}
L_w(\mathcal{M}^\uparrow)(W'W') &= \delta^\uparrow(q_0, W'W')(q_2) \\
&= \sum_{q \in Q} \delta^\uparrow(q_0, W')(q) \cdot \delta^\uparrow(q, W')(q_2) \\
&= \delta^\uparrow(q_0, W')(q_0) \cdot \delta^\uparrow(q_0, W')(q_2) + \delta^\uparrow(q_0, W')(q_1) \cdot \delta^\uparrow(q_1, W')(q_2) \\
&\quad + \delta^\uparrow(q_0, W')(q_2) \cdot \delta^\uparrow(q_2, W')(q_2) \\
&= 0.286467.
\end{aligned}$$

We now calculate $L_w(\mathcal{M}^\uparrow)(W'W')$ using Theorem 4.6. By definition, we have that

$$\begin{aligned}
\theta_{W'}(W_1) &= W'(a) \cdot \frac{W_1(a)}{W_1(a) + W_2(a)} + W'(b) \cdot \frac{W_1(b)}{W_1(b) + W_2(b)} = 0.26, \\
\theta_{W'}(W_2) &= W'(a) \cdot \frac{W_2(a)}{W_1(a) + W_2(a)} + W'(b) \cdot \frac{W_2(b)}{W_1(b) + W_2(b)} = 0.74.
\end{aligned}$$

In Example 2.4, we have obtained that $L_w(\mathcal{M})(W_1W_1) = 0.05, L_w(\mathcal{M})(W_1W_2) = 0.1975, L_w(\mathcal{M})(W_2W_1) = 0.05$, and $L_w(\mathcal{M})(W_2W_2) = 0.43$. It thus follows from Theorem 4.6 that

$$\begin{aligned}
L_w(\mathcal{M}^\uparrow)(W'W') &= L_w(\mathcal{M})(W_1W_1) \cdot \theta_{W'}(W_1) \cdot \theta_{W'}(W_1) \\
&\quad + L_w(\mathcal{M})(W_1W_2) \cdot \theta_{W'}(W_1) \cdot \theta_{W'}(W_2) \\
&\quad + L_w(\mathcal{M})(W_2W_1) \cdot \theta_{W'}(W_2) \cdot \theta_{W'}(W_1) \\
&\quad + L_w(\mathcal{M})(W_2W_2) \cdot \theta_{W'}(W_2) \cdot \theta_{W'}(W_2) \\
&= 0.286467,
\end{aligned}$$

as desired.

The proof of Theorem 4.6 proceeds along the same lines as the proof of Theorem 3.3; let us first establish the following lemma.

**Lemma 4.8.** *Let $M_w = (Q, \Sigma_w, \delta, q_0, F)$ be a PACW and $M_w^\uparrow = (Q, \mathcal{D}(\Sigma), \delta^\uparrow, q_0, F)$ be the generalized extension of $M_w$. Then for any $p, q \in Q$ and $S = W_1' \cdots W_l' \in \mathcal{D}(\Sigma)^*$, we have that*

$$\delta^\uparrow(p, S)(q) = \sum_{W_1, \ldots, W_l \in \Sigma_w} \delta(p, W_1 \cdots W_l)(q) \cdot \prod_{i=1}^{l} \theta_{W_i'}(W_i).$$

**Proof.** We prove it by induction on $l$.

1) For the basis step, namely, $l = 0$, it is trivial.

2) The induction hypothesis is that the above equation holds for $S = W_1' \cdots W_l' \in \mathcal{D}(\Sigma)^*$. We now prove the same for $SW_{l+1}'$, i.e., $W_1' \cdots W_l' W_{l+1}'$. Using the definition of $\delta^{\uparrow}$ and the induction hypothesis, we have the following computation.

$$
\begin{aligned}
\delta^{\uparrow}(p, SW'_{l+1})(q) &= \delta^{\uparrow}(p, W'_1 \cdots W'_l W'_{l+1})(q) \\
&= \Big[ \sum_{q' \in Q} \delta^{\uparrow}(p, W'_1 \cdots W'_l)(q') \cdot \delta^{\uparrow}(q', W'_{l+1}) \Big](q) \\
&= \sum_{q' \in Q} \delta^{\uparrow}(p, W'_1 \cdots W'_l)(q') \cdot \delta^{\uparrow}(q', W'_{l+1})(q) \\
&= \sum_{q' \in Q} \Big[ \sum_{W_1,\ldots,W_l \in \Sigma_w} \delta(p, W_1 \cdots W_l)(q') \cdot \prod_{i=1}^{l} \theta_{W'_i}(W_i) \Big] \cdot \delta^{\uparrow}(q', W'_{l+1})(q) \\
&= \sum_{q' \in Q} \sum_{W_1,\ldots,W_l \in \Sigma_w} \delta(p, W_1 \cdots W_l)(q') \cdot \delta^{\uparrow}(q', W'_{l+1})(q) \cdot \prod_{i=1}^{l} \theta_{W'_i}(W_i) \\
&= \sum_{q' \in Q} \sum_{W_1,\ldots,W_l \in \Sigma_w} \delta(p, W_1 \cdots W_l)(q') \cdot \Big[ \sum_{W_{l+1} \in \Sigma_w} \theta_{W'_{l+1}}(W_{l+1}) \\
&\qquad\qquad \cdot \delta(q', W_{l+1})(q) \Big] \cdot \prod_{i=1}^{l} \theta_{W'_i}(W_i) \\
&= \sum_{q' \in Q} \sum_{W_1,\ldots,W_l \in \Sigma_w} \sum_{W_{l+1} \in \Sigma_w} \delta(p, W_1 \cdots W_l)(q') \cdot \delta(q', W_{l+1})(q) \\
&\qquad\qquad \cdot \prod_{i=1}^{l+1} \theta_{W'_i}(W_i) \\
&= \sum_{q' \in Q} \sum_{W_1,\ldots,W_{l+1} \in \Sigma_w} \delta(p, W_1 \cdots W_l)(q') \cdot \delta(q', W_{l+1})(q) \cdot \prod_{i=1}^{l+1} \theta_{W'_i}(W_i) \\
&= \sum_{W_1,\ldots,W_{l+1} \in \Sigma_w} \sum_{q' \in Q} \delta(p, W_1 \cdots W_l)(q') \cdot \delta(q', W_{l+1})(q) \cdot \prod_{i=1}^{l+1} \theta_{W'_i}(W_i) \\
&= \sum_{W_1,\ldots,W_{l+1} \in \Sigma_w} \Big[ \sum_{q' \in Q} \delta(p, W_1 \cdots W_l)(q') \cdot \delta(q', W_{l+1})(q) \Big] \cdot \prod_{i=1}^{l+1} \theta_{W'_i}(W_i) \\
&= \sum_{W_1,\ldots,W_{l+1} \in \Sigma_w} \delta(p, W_1 \cdots W_{l+1})(q) \cdot \prod_{i=1}^{l+1} \theta_{W'_i}(W_i),
\end{aligned}
$$

i.e., $\delta^{\uparrow}(p, SW'_{l+1})(q) = \sum_{W_1,\ldots,W_{l+1} \in \Sigma_w} \delta(p, W_1 \cdots W_{l+1})(q) \cdot \prod_{i=1}^{l+1} \theta_{W'_i}(W_i)$. This finishes the proof of the lemma. □

Based on the above lemma, the proof of Theorem 4.6 is straightforward.

**Proof of Theorem 4.6.** By the definition of word languages and Lemma 4.8, we have the following equation.

$$\begin{aligned}
L_w(M_w^\uparrow)(S) &= \sum_{q \in F} \delta^\uparrow(q_0, W_1' \cdots W_l')(q) \\
&= \sum_{q \in F} \sum_{W_1,\ldots,W_l \in \Sigma_w} \delta(q_0, W_1 \cdots W_l)(q) \cdot \prod_{i=1}^{l} \theta_{W_i'}(W_i) \\
&= \sum_{W_1,\ldots,W_l \in \Sigma_w} \sum_{q \in F} \delta(q_0, W_1 \cdots W_l)(q) \cdot \prod_{i=1}^{l} \theta_{W_i'}(W_i) \\
&= \sum_{W_1,\ldots,W_l \in \Sigma_w} \left[\sum_{q \in F} \delta(q_0, W_1 \cdots W_l)(q)\right] \cdot \prod_{i=1}^{l} \theta_{W_i'}(W_i) \\
&= \sum_{W_1,\ldots,W_l \in \Sigma_w} L_w(M_w)(W_1 \cdots W_l) \cdot \prod_{i=1}^{l} \theta_{W_i'}(W_i),
\end{aligned}$$

which proves the theorem. $\square$

After discussing the generalized extensions of PACWs under the probability interpretation of words, we point out some slight modifications for the case of the possibility interpretation of words.

**Remark 4.9.** If the words in this subsection are interpreted as possibility distributions, then to make the definition of generalized extensions true, we have to modify Definition 4.1 by substituting $\|W'\|(\sigma)$ for $W'(\sigma)$, where $\|W'\|(\sigma)$ stands for $W'(\sigma)/\sum_{\tau \in \Sigma} W'(\tau)$. It is not hard to check that this substitution is prerequisite for $\delta^\uparrow(p, W') \in \mathcal{D}(Q)$. Correspondingly, the notation $\theta_{W'}(W)$ denoting $\sum_{\sigma \in \Sigma} W'(\sigma) \cdot \chi_\sigma(W)$ is replaced by $\tilde{\theta}_{W'}(W) = \sum_{\sigma \in \Sigma} \|W'\|(\sigma) \cdot \chi_\sigma(W)$. In addition, since we are interpreting words as possibility distributions, it is natural to require that $\Sigma_w \subseteq \mathcal{F}(\Sigma)$ and replace $\mathcal{D}(\Sigma)$ by $\mathcal{F}(\Sigma)$. With these substitutions, all of the results in this subsection hold, except for Proposition 4.4 and Corollary 4.5 which need more modifications. For example, Definition 4.1 can be stated as follows.

**Definition 4.1'.** Let $M_w = (Q, \Sigma_w, \delta, q_0, F)$ be a PACW with $\Sigma_w \subseteq \mathcal{F}(\Sigma)$. The *generalized extension* of $M_w$ is a PACAW $M_w^\uparrow = (Q, \mathcal{F}(\Sigma), \delta^\uparrow, q_0, F)$, where:

(a) The components $Q, q_0, F$ are the same as those of $M_w$.

(b) $\mathcal{F}(\Sigma)$ consists of all possibility distributions (i.e., fuzzy subsets) over the underlying input alphabet of $M_w$.

(c) $\delta^\uparrow$, the transition probability function, is a mapping from $Q \times \mathcal{F}(\Sigma)$ to $\mathcal{D}(Q)$ defined by

$$\delta^\uparrow(p, W') = \sum_{W \in \Sigma_w} \tilde{\theta}_{W'}(W) \cdot \delta(p, W)$$

for any $(p, W') \in Q \times \mathcal{F}(\Sigma)$, where $\tilde{\theta}_{W'}(W) = \sum_{\sigma \in \Sigma} \|W'\|(\sigma) \cdot [W(\sigma)/\sum_{U \in \Sigma_w} U(\sigma)]$.

In terms of possibility distributions, Proposition 4.4 and Corollary 4.5 can be stated as follows.

**Proposition 4.4'.** *Suppose that the linear combination $W' = \sum_{i=1}^{l} k_i W'_i$ with $W'_i \in \mathcal{F}(\Sigma)$ is a word in $\mathcal{F}(\Sigma)$. Then*

$$\delta^{\uparrow}(p, W') = \frac{1}{\sum_{\sigma \in \Sigma} W'(\sigma)} \sum_{i=1}^{l} k_i \cdot \delta^{\uparrow}(p, W'_i).$$

**Corollary 4.5'.** *To compute $\delta^{\uparrow}(p, W')$ for any $(p, W') \in Q \times \mathcal{F}(\Sigma)$, we can follow the following steps:*

(1) *Find any $n$ linearly independent words in $\mathcal{F}(\Sigma)$, say $W'_1, \ldots, W'_n$, and write $W' = k_1 W'_1 + \cdots + k_n W'_n$.*

(2) *Compute $\delta^{\uparrow}(p, W'_i)$ for all $i = 1, \ldots, n$.*

(3) *Set $k'_i = k_i / \sum_{\sigma \in \Sigma} W'(\sigma)$, and then compute the sum $\sum_{i=1}^{n} k'_i \cdot \delta^{\uparrow}(p, W'_i)$ which exactly equals $\delta^{\uparrow}(p, W')$.*

### 4.2 Analytical properties of the generalized extensions

In this subsection, we pay attention to two analytical properties of the generalized extensions. Intuitively, the analytical properties show us that the transition probabilities of two near inputs are near as well.

Let us keep the assumption that $\Sigma = \{\sigma_1, \sigma_2, \ldots, \sigma_n\}$. Recall that each word $W \in \mathcal{D}(\Sigma)$ can be identified with an $n$-dimensional stochastic row vector $[W(\sigma_1), W(\sigma_2), \ldots, W(\sigma_n)]$. Since such a stochastic row vector corresponds to a point in the $n$-dimensional Euclidean space $\mathbb{R}^n$ (in fact, $\mathcal{D}(\Sigma)$ is exactly the $(n-1)$-dimensional simplex), we can discuss the distance between two words in $\mathcal{D}(\Sigma)$. More explicitly, for any two words $W'$ and $W''$ in $\mathcal{D}(\Sigma)$, it follows from the definition of Euclidean metric that the distance $d(W', W'')$ is given by

$$d(W', W'') = \sqrt{\sum_{i=1}^{n} (k'_i - k''_i)^2},$$

where $k'_i = W'(\sigma_i)$ and $k''_i = W''(\sigma_i)$.

The result below shows that if two words are near enough, then the transition probabilities when inputting them at the same state are nearby. In this subsection, we abuse the notation $\epsilon$ and $\delta$ used in PACWs as well as in the following epsilon-delta language, and also abuse the notation "$|\ \ |$" to denote the cardinality of a set and the absolute value of a real number.

**Proposition 4.10.** *For any $p, q \in Q$, the function $\delta^{\uparrow}(p, x)(q)$ is uniformly continuous on $\mathcal{D}(\Sigma)$. In other words, for any $\epsilon > 0$, there exists a $\delta > 0$ such that $|\delta^{\uparrow}(p, W')(q) - \delta^{\uparrow}(p, W'')(q)| < \epsilon$ for any $p, q \in Q$, whenever $d(W', W'') < \delta$ with $W', W'' \in \mathcal{D}(\Sigma)$.*

**Proof.** Suppose that $W'(\sigma_i) = k'_i$ and $W''(\sigma_i) = k''_i$ for $i = 1, \ldots, n$. Taking $\delta = \epsilon/\sqrt{n}$, we see by definition that $\sqrt{\sum_{i=1}^{n}(k'_i - k''_i)^2} < \epsilon/\sqrt{n}$. Moreover, using the equation (2) we have that

$$
\begin{aligned}
\left|\delta^{\uparrow}(p, W')(q) - \delta^{\uparrow}(p, W'')(q)\right| &= \left|\sum_{i=1}^{n} k'_i \cdot \delta^{\uparrow}(p, \widehat{\sigma}_i)(q) - \sum_{i=1}^{n} k''_i \cdot \delta^{\uparrow}(p, \widehat{\sigma}_i)(q)\right| \\
&= \left|\sum_{i=1}^{n} (k'_i - k''_i) \cdot \delta^{\uparrow}(p, \widehat{\sigma}_i)(q)\right| \\
&\leq \sum_{i=1}^{n} \left|(k'_i - k''_i) \cdot \delta^{\uparrow}(p, \widehat{\sigma}_i)(q)\right| \\
&= \sum_{i=1}^{n} |k'_i - k''_i| \cdot \delta^{\uparrow}(p, \widehat{\sigma}_i)(q) \\
&\leq \sum_{i=1}^{n} |k'_i - k''_i|.
\end{aligned}
$$

Note that the arithmetic mean of a set of values is not greater than the root-mean-square of these values, i.e., $\left|(\sum_{i=1}^{n} a_i)/n\right| \leq \sqrt{(\sum_{i=1}^{n} a_i^2)/n}$ for any $a_1, \ldots, a_n \in \mathbb{R}$. We thus get that

$$
\begin{aligned}
\sum_{i=1}^{n} |k'_i - k''_i| &\leq \sqrt{n} \cdot \sqrt{\sum_{i=1}^{n} (k'_i - k''_i)^2} \\
&< \sqrt{n} \cdot \frac{\epsilon}{\sqrt{n}} = \epsilon,
\end{aligned}
$$

that is, $\left|\delta^{\uparrow}(p, W')(q) - \delta^{\uparrow}(p, W'')(q)\right| < \epsilon$, finishing the proof. □

To give an illustration, let us examine the transition probability function of $L_w(\mathcal{M}^{\uparrow})$ obtained in Example 4.2.

**Example 4.11.** Keep the data in Example 4.7, where we have obtained all transition probabilities $\delta^{\uparrow}(p, W')(q)$ for $W' = 0.2\backslash a + 0.8\backslash b$ and any $p, q \in \{q_0, q_1, q_2\}$; for instance,

$$\delta^{\uparrow}(q_1, W') = 0.141\backslash q_0 + 0.55\backslash q_1 + 0.309\backslash q_2.$$

Let us consider which words $W'' \in \mathcal{D}(\{a, b\})$ can be inputted at any state $p \in \{q_0, q_1, q_2\}$ such that $|\delta^{\uparrow}(p, W')(q) - \delta^{\uparrow}(p, W'')(q)| < 0.001$ for all $q \in \{q_0, q_1, q_2\}$. To this end, we take $\epsilon = 0.001$ and apply Proposition 4.10. Let $W'' = \alpha\backslash a + (1-\alpha)\backslash b \in \mathcal{D}(\{a, b\})$. By the proof of Proposition 4.10, taking $\delta = \sqrt{2}/2000$, we can get that $|\delta^{\uparrow}(p, W')(q) - \delta^{\uparrow}(p, W'')(q)| < 0.001$, whenever $d(W', W'') < \delta$. The latter is equivalent to $\alpha \in (0.1995, 0.2005)$ by a routine computation. Summarly, for any $W'' = \alpha\backslash a + (1-\alpha)\backslash b$ with $\alpha \in (0.1995, 0.2005)$, we have that $|\delta^{\uparrow}(p, W')(q) - \delta^{\uparrow}(p, W'')(q)| < 0.001$ for any $p, q \in \{q_0, q_1, q_2\}$.

More concretely, take $\alpha = 0.2004 \in (0.1995, 0.2005)$ and $p = q_1$ as an example. It follows from the formula in Example 4.2 that

$$
\begin{aligned}
\delta^{\uparrow}(q_1, W'') &= (0.085 + 0.28\alpha)\backslash q_0 + 0.55\backslash q_1 + (0.365 - 0.28\alpha)\backslash q_2 \\
&= 0.141112\backslash q_0 + 0.55\backslash q_1 + 0.308888\backslash q_2.
\end{aligned}
$$

As a result, we see that

$$|\delta^\uparrow(q_1, W')(q_0) - \delta^\uparrow(q_1, W'')(q_0)| \;=\; |0.141 - 0.141112| = 0.000112 < 0.001,$$

$$|\delta^\uparrow(q_1, W')(q_1) - \delta^\uparrow(q_1, W'')(q_1)| \;=\; |0.55 - 0.55| = 0 < 0.001,$$

$$|\delta^\uparrow(q_1, W')(q_2) - \delta^\uparrow(q_1, W'')(q_2)| \;=\; |0.309 - 0.308888| = 0.000112 < 0.001,$$

as desired.

We further compare the accepting probabilities of two strings of near words. As expected, if every corresponding components of two strings are near enough, then the accepting probabilities of the two strings are nearby. To state this in a mathematical term, we need a metric space which makes the accepting probabilities of two strings comparable.

Notice that the set $\mathcal{D}(\Sigma)$ with the Euclidean metric $d$ gives rise to a metric space. For any $l \geq 1$, denote by $\mathcal{D}(\Sigma)^l$ the Cartesian product of $l$ copies of $\mathcal{D}(\Sigma)$; any element $(W'_1, \ldots, W'_l) \in \mathcal{D}(\Sigma)^l$ will be written as $W'_1 \cdots W'_l$. We define a function

$$\begin{array}{rcl} d_l : \quad \mathcal{D}(\Sigma)^l \times \mathcal{D}(\Sigma)^l & \longrightarrow & \mathbb{R} \\ (W'_1 \cdots W'_l, W''_1 \cdots W''_l) & \longmapsto & \max_{i=1}^{l} d(W'_i, W''_i). \end{array}$$

It is easy to check that $d_l$ is a metric on $\mathcal{D}(\Sigma)^l$, which makes $\mathcal{D}(\Sigma)^l$ into a metric space. Observe that for any given $M_w^\uparrow$, the word language of $M_w^\uparrow$ gives a function $L_w(M_w^\uparrow)|_{\mathcal{D}(\Sigma)^l} : \mathcal{D}(\Sigma)^l \longrightarrow \mathbb{R}$ that maps $W'_1 \cdots W'_l$ to $L_w(M_w^\uparrow)(W'_1 \cdots W'_l)$. Such a function has a good property, as shown below.

**Proposition 4.12.** *For any $l \geq 1$, the function $L_w(M_w^\uparrow)|_{\mathcal{D}(\Sigma)^l}$ is uniformly continuous on $\mathcal{D}(\Sigma)^l$. In other words, for any $\epsilon > 0$, there exists a $\delta > 0$ such that $\left|L_w(M_w^\uparrow)(W'_1 \cdots W'_l) - L_w(M_w^\uparrow)(W''_1 \cdots W''_l)\right| < \epsilon$, whenever $d(W'_i, W''_i) < \delta$ holds for every pair $W'_i, W''_i \in \mathcal{D}(\Sigma)$, where $1 \leq i \leq l$.*

**Proof.** What we actually prove first, by induction on $l$, is the following claim: for any $\epsilon > 0$, there exists a $\delta > 0$ such that $\left|\delta^\uparrow(p, W'_1 \cdots W'_l)(q) - \delta^\uparrow(p, W''_1 \cdots W''_l)(q)\right| < \epsilon$ for any $p, q \in Q$, whenever $d(W'_i, W''_i) < \delta$ holds for every pair $W'_i, W''_i \in \mathcal{D}(\Sigma)$, where $1 \leq i \leq l$.
1) The case of $l = 1$ follows immediately from Proposition 4.10.
2) Assume that the statement holds for the case of $l - 1$. We now consider the case of $l$. Given any $\epsilon > 0$, it follows from the induction assumption that for $\epsilon/(2|Q|)$, there exists a $\delta' > 0$ such that $\left|\delta^\uparrow(p, W'_1 \cdots W'_{l-1})(q) - \delta^\uparrow(p, W''_1 \cdots W''_{l-1})(q)\right| < \epsilon/(2|Q|)$ for any $p, q \in Q$, whenever $d(W'_i, W''_i) < \delta'$ holds for every pair $W'_i, W''_i \in \mathcal{D}(\Sigma)$, where $1 \leq i \leq l-1$. It also follows from the basis step or Proposition 4.10 that for the given $\epsilon$, there exists a $\delta'' > 0$ such that $\left|\delta^\uparrow(p, W'_l)(q) - \delta^\uparrow(p, W''_l)(q)\right| < \epsilon/(2|Q|)$ for any $p, q \in Q$, whenever $d(W'_l, W''_l) < \delta''$. Taking $\delta = \min\{\delta', \delta''\}$, we get that

$$\left|\delta^\uparrow(p, W'_1 \cdots W'_{l-1} W'_l)(q) - \delta^\uparrow(p, W''_1 \cdots W''_{l-1} W''_l)(q)\right|$$
$$= \left|\sum_{q' \in Q} \delta^\uparrow(p, W'_1 \cdots W'_{l-1})(q') \cdot \delta^\uparrow(q', W'_l)(q) - \sum_{q' \in Q} \delta^\uparrow(p, W''_1 \cdots W''_{l-1})(q') \cdot \delta^\uparrow(q', W''_l)(q)\right|$$
$$= \left|\sum_{q' \in Q} \left[\delta^\uparrow(p, W'_1 \cdots W'_{l-1})(q') \cdot \delta^\uparrow(q', W'_l)(q) - \delta^\uparrow(p, W''_1 \cdots W''_{l-1})(q') \cdot \delta^\uparrow(q', W''_l)(q)\right]\right|$$
$$\leq \sum_{q' \in Q} \left|\delta^\uparrow(p, W'_1 \cdots W'_{l-1})(q') \cdot \delta^\uparrow(q', W'_l)(q) - \delta^\uparrow(p, W''_1 \cdots W''_{l-1})(q') \cdot \delta^\uparrow(q', W''_l)(q)\right|$$
$$= \sum_{q' \in Q} \left|\delta^\uparrow(q', W'_l)(q) \cdot \left[\delta^\uparrow(p, W'_1 \cdots W'_{l-1})(q') - \delta^\uparrow(p, W''_1 \cdots W''_{l-1})(q')\right]\right.$$
$$\left. + \delta^\uparrow(p, W''_1 \cdots W''_{l-1})(q') \cdot \left[\delta^\uparrow(q', W'_l)(q) - \delta^\uparrow(q', W''_l)(q)\right]\right|$$
$$\leq \sum_{q' \in Q} \left\{\left|\delta^\uparrow(q', W'_l)(q) \cdot \left[\delta^\uparrow(p, W'_1 \cdots W'_{l-1})(q') - \delta^\uparrow(p, W''_1 \cdots W''_{l-1})(q')\right]\right|\right.$$
$$\left. + \left|\delta^\uparrow(p, W''_1 \cdots W''_{l-1})(q') \cdot \left[\delta^\uparrow(q', W'_l)(q) - \delta^\uparrow(q', W''_l)(q)\right]\right|\right\}$$
$$\leq \sum_{q' \in Q} \left\{\left|\delta^\uparrow(p, W'_1 \cdots W'_{l-1})(q') - \delta^\uparrow(p, W''_1 \cdots W''_{l-1})(q')\right|\right.$$
$$\left. + \left|\delta^\uparrow(q', W'_l)(q) - \delta^\uparrow(q', W''_l)(q)\right|\right\}$$
$$< |Q| \cdot \left(\frac{\epsilon}{2|Q|} + \frac{\epsilon}{2|Q|}\right) = \epsilon,$$

namely, $\left|\delta^\uparrow(p, W'_1 \cdots W'_{l-1} W'_l)(q) - \delta^\uparrow(p, W''_1 \cdots W''_{l-1} W''_l)(q)\right| < \epsilon$. This completes the proof of the claim.

Now, let us use the claim and the definition of word languages to prove the proposition. For any $\epsilon > 0$, it follows from the claim that there exists a $\delta > 0$ such that $\left|\delta^\uparrow(q_0, W'_1 \cdots W'_l)(q) - \delta^\uparrow(q_0, W''_1 \cdots W''_l)(q)\right| < \epsilon/|F|$ for any $q \in F$, whenever $d(W'_i, W''_i) < \delta$ holds for every pair $W'_i, W''_i \in \mathcal{D}(\Sigma)$, where $1 \leq i \leq l$. By the definition of word languages, we have that

$$\left|L_w(M_w^\uparrow)(W'_1 \cdots W'_l) - L_w(M_w^\uparrow)(W''_1 \cdots W''_l)\right|$$
$$= \left|\sum_{q \in F} \delta^\uparrow(q_0, W'_1 \cdots W'_l)(q) - \sum_{q \in F} \delta^\uparrow(q_0, W''_1 \cdots W''_l)(q)\right|$$
$$= \left|\sum_{q \in F} \left[\delta^\uparrow(q_0, W'_1 \cdots W'_l)(q) - \delta^\uparrow(q_0, W''_1 \cdots W''_l)(q)\right]\right|$$
$$\leq \sum_{q \in F} \left|\delta^\uparrow(q_0, W'_1 \cdots W'_l)(q) - \delta^\uparrow(q_0, W''_1 \cdots W''_l)(q)\right|$$
$$< |F| \cdot \frac{\epsilon}{|F|} = \epsilon,$$

which finishes the proof of the proposition. □

**Remark 4.13.** When the words are interpreted as possibility distributions, one can readily show that Proposition 4.10 and Proposition 4.12 remain valid by replacing $\mathcal{D}(\Sigma)$ with $\mathcal{F}(\Sigma)$.

## 5 Retractions and generalized extensions of PGCWs

In Sections 3 and 4, we introduced the concepts of retractions and generalized extensions of PACWs. In fact, these concepts are appropriate for PGCWs. We briefly discuss them in this section.

Motivated by the retractions and the generalized extensions of PACWs, we can directly define the retractions and the generalized extensions of PGCWs as follows.

**Definition 5.1.** Let $G_w = (V, \Sigma_w, P_w, S)$ be a PGCW. The *retraction* of $G_w$ is a probabilistic grammar $G_w^\downarrow = (V, \Sigma, P, S)$, where $\Sigma$ is the underlying terminal set of $\Sigma_w$ and

$$P = \{A \to aB : \exists W \in \Sigma_w \text{ such that } A \to WB \in P_w \text{ and } W(a) \neq 0\} \cup \{A \to \epsilon : A \in V\}$$

with

$$Pr^\downarrow(A \to aB) = \sum_{W \in \Sigma_w} \frac{W(a)}{\sum_{U \in \Sigma_w} U(a)} Pr(A \to WB),$$
$$Pr^\downarrow(A \to \epsilon) = Pr(A \to \epsilon).$$

Analogous, we have the following definition.

**Definition 5.2.** The *generalized extension* of a PGCW $G_w = (V, \Sigma_w, P_w, S)$ is a probabilistic grammar for computing with all words (PGCAW), denoted $G_w^\uparrow = (V, \mathcal{D}(\Sigma), P, S)$, where $\mathcal{D}(\Sigma)$ consists of all probabilistic distributions over the underlying terminal set of $\Sigma_w$ and

$$P = \{A \to W'B : A, B \in V,\ W' \in \mathcal{D}(\Sigma)\} \cup \{A \to \epsilon : A \in V\}$$

with

$$Pr^\uparrow(A \to W'B) = \sum_{W \in \Sigma_w} \theta_{W'}(W) \cdot Pr(A \to WB),$$
$$Pr^\uparrow(A \to \epsilon) = Pr(A \to \epsilon).$$

In the above, $\theta_{W'}(W) = \sum_{a \in \Sigma} W'(a) \cdot \left[W(a)/\sum_{U \in \Sigma_w} U(a)\right]$, which is the same as in Definition 4.1.

In terms of PGCWs, there is a retraction principle from computing with words to computing with values.

**Theorem 5.3.** *Suppose that $G_w = (V, \Sigma_w, P_w, S)$ is a PGCW and $G_w^\downarrow = (V, \Sigma, P, S)$ is the retraction of $G_w$. Then for any $s = a_1 \cdots a_l \in \Sigma^*$, we have that*

$$L(G_w^\downarrow)(s) = \sum_{W_1, \ldots, W_l \in \Sigma_w} L_w(G_w)(W_1 \cdots W_l) \cdot \prod_{i=1}^{l} \chi_{a_i}(W_i),$$

*where $\chi_{a_i}(W_i) = W_i(a_i)/\sum_{U \in \Sigma_w} U(a_i)$.*

The generalized extension principle for PGCWs can be stated as follows.

**Theorem 5.4.** *Suppose that $G_w = (V, \Sigma_w, P_w, S)$ is a PGCW and $G_w^\uparrow = (V, \mathcal{D}(\Sigma), P, S)$ is the generalized extension of $G_w$. Then for any $s = W_1' \cdots W_l' \in \mathcal{D}(\Sigma)^*$, we have that*

$$L_w(G_w^\uparrow)(s) = \sum_{W_1, \ldots, W_l \in \Sigma_w} L_w(G_w)(W_1 \cdots W_l) \cdot \prod_{i=1}^{l} \theta_{W_i'}(W_i),$$

*where $\theta_{W_i'}(W_i) = \sum_{a \in \Sigma} W_i'(a) \cdot \big[W_i(a) / \sum_{U \in \Sigma_w} U(a)\big]$.*

The proofs of Theorems 5.3 and 5.4 follow immediately from Theorem 3.3, Theorem 4.6, and the results obtained in the next section, so we leave them to the end of Section 6.

## 6  Retractions and generalized extensions are equivalence-preserving

As mentioned in Introduction, there are three kinds of equivalences among PACVs and probabilistic grammars: the equivalence between two PACVs, the equivalence between two probabilistic grammars, and the equivalence between PACVs and probabilistic grammars; see Figure 3. In this section, we examine the preservation of these equivalences under retractions and generalized extensions.

Let us start with the following definition.

**Definition 6.1.** *Two PACVs $M' = (Q', \Sigma, \delta', q_0', F')$ and $M'' = (Q'', \Sigma, \delta'', q_0'', F'')$ are equivalent if $L(M')(s) = L(M'')(s)$ for all strings $s \in \Sigma^*$; two probabilistic grammars $G' = (V', \Sigma, P', S')$ and $G'' = (V'', \Sigma, P'', S'')$ are equivalent if $L(G')(s) = L(G'')(s)$ for all strings $s \in \Sigma^*$.*

Recall that the PACV $M'$ is said to be equivalent to the probabilistic grammar $G'$, if $L(M')(s) = L(G')(s)$ for all strings $s \in \Sigma^*$. Clearly, these definitions are applicable to PACWs and PACAWs (and also PGCWs and PGCAWs) by replacing $\Sigma$ with $\Sigma_w$ and $\mathcal{D}(\Sigma)$, respectively. The next proposition shows us that if two PACWs are equivalent, then so are their retractions and generalized extensions; this corresponds to the two left boxes in Figure 3.

**Proposition 6.2.** *Suppose that $M_w' = (Q', \Sigma_w, \delta', q_0', F')$ and $M_w'' = (Q'', \Sigma_w, \delta'', q_0'', F'')$ are two equivalent PACWs. Then:*

*(a) $M_w'^\downarrow = (Q', \Sigma, \delta'^\downarrow, q_0', F')$ and $M_w''^\downarrow = (Q'', \Sigma, \delta''^\downarrow, q_0'', F'')$ are equivalent.*

*(b) $M_w'^\uparrow = (Q', \mathcal{D}(\Sigma), \delta'^\uparrow, q_0', F')$ and $M_w''^\uparrow = (Q'', \mathcal{D}(\Sigma), \delta''^\uparrow, q_0'', F'')$ are equivalent.*

**Proof.** The proofs of $(a)$ and $(b)$ are similar, so we only prove the assertion $(a)$. The hypothesis means that $L_w(M_w')(W_1 \cdots W_l) = L_w(M_w'')(W_1 \cdots W_l)$ for all $W_1 \cdots W_l \in \Sigma_w^*$. To show that $M_w'^\downarrow$ and $M_w''^\downarrow$ are equivalent, by Definition 6.1 we need to verify that $L(M_w'^\downarrow)(s) = L(M_w''^\downarrow)(s)$ for any $s = \sigma_1 \cdots \sigma_l \in \Sigma^*$. It follows from Theorem 3.3 and the

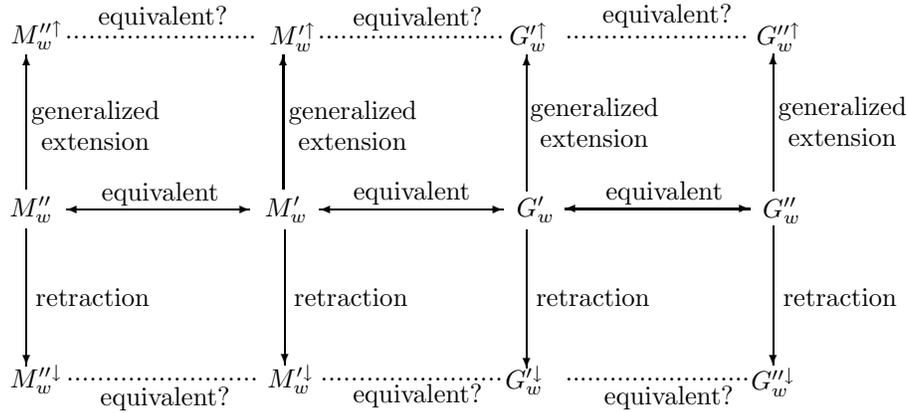

Figure 3: Equivalences under retractions and generalized extensions.

hypothesis that

$$\begin{aligned} L(M_w'^{\downarrow})(s) &= \sum_{W_1,\ldots,W_l \in \Sigma_w} L_w(M_w')(W_1 \cdots W_l) \cdot \prod_{i=1}^{l} \chi_{\sigma_i}(W_i) \\ &= \sum_{W_1,\ldots,W_l \in \Sigma_w} L_w(M_w'')(W_1 \cdots W_l) \cdot \prod_{i=1}^{l} \chi_{\sigma_i}(W_i) \\ &= L(M_w''^{\downarrow})(s), \end{aligned}$$

as desired. This finishes the proof of the proposition. □

We continue to discuss the preservation of equivalence between PACWs and PGCWs, which corresponds to the two in-between boxes in Figure 3.

**Proposition 6.3.** *Let $M_w' = (Q', \Sigma_w, \delta', q_0', F')$ be a PACW and $G_w' = (V', \Sigma_w, P_w', S')$ be a PGCW. If $M_w'$ is equivalent to $G_w'$, then:*

(a) $M_w'^{\downarrow} = (Q', \Sigma, \delta'^{\downarrow}, q_0', F')$ and $G_w'^{\downarrow} = (V', \Sigma, P', S')$ are equivalent.

(b) $M_w'^{\uparrow} = (Q', \mathcal{D}(\Sigma), \delta'^{\uparrow}, q_0', F')$ and $G_w'^{\uparrow} = (V', \mathcal{D}(\Sigma), P', S')$ are equivalent.

**Proof.** We only prove $(a)$; the part $(b)$ can be proved in a similar way. Using the construction in Section 2.3, we have the PACW $M_1 = (V', \Sigma_w, \delta_1, S', F_1)$ induced from $G_w'$, where $F_1 = \{A' \in V' : Pr'(A' \to \epsilon) = 1\}$ and $\delta_1$ is defined by $\delta_1(A', W)(B') = Pr'(A' \to WB')$ for any $A', B' \in V'$ and $W \in \Sigma_w$. Therefore, $M_w'$ and $M_1$ are equivalent. We thus get by Proposition 6.2 that $M_w'^{\downarrow}$ and $M_1^{\downarrow}$ are equivalent. Consequently, to prove $(a)$, it is sufficient to show that $M_1^{\downarrow}$ and $G_w'^{\downarrow}$ are equivalent. We verify it by showing that $M_1^{\downarrow}$ is equivalent to the PACV, say $M_2$, induced from $G_w'^{\downarrow}$.

By definition, $M_1^{\downarrow} = (V', \Sigma, \delta_1^{\downarrow}, S', F_1)$, where

$$\begin{aligned} \delta_1^{\downarrow}(A', a)(B') &= \sum_{W \in \Sigma_w} \frac{W(a)}{\sum_{U \in \Sigma_w} U(a)} \cdot \delta_1(A', W)(B') \\ &= \sum_{W \in \Sigma_w} \frac{W(a)}{\sum_{U \in \Sigma_w} U(a)} Pr'(A' \to WB') \end{aligned}$$

for any $A', B' \in V'$ and $a \in \Sigma$. By the definition of retractions, we see that in $G_w'^{\downarrow}$,

$$P' = \{A' \to aB' : \exists W \in \Sigma_w \text{ such that } A' \to WB' \in P_w' \text{ and } W(a) \neq 0\}$$
$$\cup \{A' \to \epsilon : A' \in V'\}$$

with

$$Pr'^{\downarrow}(A' \to aB') = \sum_{W \in \Sigma_w} \frac{W(a)}{\sum_{U \in \Sigma_w} U(a)} Pr'(A' \to WB'),$$
$$Pr'^{\downarrow}(A' \to \epsilon) = Pr'(A' \to \epsilon).$$

Again, using the construction in Section 2.3, we have the PACV $M_2 = (V', \Sigma, \delta_2, S', F_2)$ induced from $G_w'^{\downarrow}$, where

$$\begin{aligned} F_2 &= \{A' \in V' : Pr'^{\downarrow}(A' \to \epsilon) = 1\} \\ &= \{A' \in V' : Pr'(A' \to \epsilon) = 1\} \end{aligned}$$

and $\delta_2$ is defined by

$$\begin{aligned} \delta_2(A', a)(B') &= Pr'^{\downarrow}(A' \to aB') \\ &= \sum_{W \in \Sigma_w} \frac{W(a)}{\sum_{U \in \Sigma_w} U(a)} Pr'(A' \to WB') \end{aligned}$$

for any $A', B' \in V'$ and $a \in \Sigma$. Whence, $F_2 = F_1$ and $\delta_2 = \delta_1^{\downarrow}$. So we get that $M_1^{\downarrow} = M_2$, finishing the proof of the part $(a)$. $\square$

Based on the above result, it is easy to show that if two PGCWs are equivalent, then so are their retractions and generalized extensions; this corresponds to the two right boxes in Figure 3.

**Proposition 6.4.** *Suppose that $G_w' = (V', \Sigma_w, P', S')$ and $G_w'' = (V'', \Sigma_w, P_w'', S'')$ are two equivalent PGCWs. Then:*

(a) $G_w'^{\downarrow} = (V', \Sigma, P', S')$ *and* $G_w''^{\downarrow} = (V'', \Sigma, P'', S'')$ *are equivalent.*

(b) $G_w'^{\uparrow} = (V', \mathcal{D}(\Sigma), P', S')$ *and* $G_w''^{\uparrow} = (V'', \mathcal{D}(\Sigma), P'', S'')$ *are equivalent.*

**Proof.** We only prove $(a)$. The proof of $(b)$ goes along the same lines as that of $(a)$, so we omit it. Let $M_w$ be the PACW induced from $G_w'$. It follows that $M_w$ and $G_w'$ are equivalent, and also $M_w$ and $G_w''$ are equivalent. By Proposition 6.3, we see that $M_w^{\downarrow}$ and $G_w'^{\downarrow}$ are equivalent, and also $M_w^{\downarrow}$ and $G_w''^{\downarrow}$ are equivalent. Therefore, $G_w'^{\downarrow}$ and $G_w''^{\downarrow}$ are equivalent, as desired. $\square$

**Remark 6.5.** When the words are interpreted as possibility distributions, it is not hard to show that the results given in this section remain true by replacing $\mathcal{D}(\Sigma)$ with $\mathcal{F}(\Sigma)$.

We end this section with the proofs of Theorems 5.3 and 5.4.

**Proof of Theorem 5.3.** Let $M_w$ be the PACW induced from $G_w$. Then $G_w$ and $M_w$ are equivalent. By Proposition 6.3, we see that $G_w^{\downarrow}$ and $M_w^{\downarrow}$ are equivalent.

Consequently, it follows from Theorem 3.3 that for any $s = a_1 \cdots a_l \in \Sigma^*$,

$$
\begin{aligned}
L(G_w^\downarrow)(s) &= L(M_w^\downarrow)(s) \\
&= \sum_{W_1,\ldots,W_l \in \Sigma_w} L_w(M_w)(W_1 \cdots W_l) \cdot \prod_{i=1}^l \chi_{a_i}(W_i) \\
&= \sum_{W_1,\ldots,W_l \in \Sigma_w} L_w(G_w)(W_1 \cdots W_l) \cdot \prod_{i=1}^l \chi_{a_i}(W_i).
\end{aligned}
$$

This completes the proof. $\square$

Theorem 5.4 can be easily proved by using Theorem 4.6 and following along the same lines as that of Theorem 5.3.

## 7 Relationships among retractions, extensions, and generalized extensions

So far, we have seen three kinds of transformations among PACVs, PACWs, and PACAWs (correspondingly, three kinds of transformations among probabilistic grammars, PGCWs, and PGCAWs), that is, the extensions in Definition 2.2, the retractions, and the generalized extensions. In fact, they are related, and we now provide some relationships in this section.

We first show that the extension given by Definition 2.2 is a special case of the generalized extension introduced in Section 4.1. To see this, we only need to regard a PACV as a PACW by identifying an input $\sigma$ with the Dirac distribution $\hat{\sigma}$ for $\sigma$. More explicitly, for a given PACV $M_v = (Q, \Sigma, \delta_v, q_0, F)$, we identify $M_v$ with a special PACW $M_w = (Q, \Sigma_w, \delta_w, q_0, F)$, where two different components are

$$
\begin{aligned}
\Sigma_w &= \{\hat{\sigma} : \hat{\sigma} \text{ is the Dirac distribution for } \sigma \in \Sigma\} \text{ and} \\
\delta_w(p, \hat{\sigma}) &= \delta_v(p, \sigma) \text{ for any } (p, \hat{\sigma}) \in Q \times \Sigma_w.
\end{aligned}
$$

It follows from (1) in Definition 3.1 that

$$
\begin{aligned}
\delta_w^\downarrow(p, \sigma) &= \sum_{\hat{\tau} \in \Sigma_w} \frac{\hat{\tau}(\sigma)}{\sum_{\hat{\pi} \in \Sigma_w} \hat{\pi}(\sigma)} \cdot \delta_w(p, \hat{\tau}) \\
&= \frac{\hat{\sigma}(\sigma)}{\hat{\sigma}(\sigma)} \cdot \delta_w(p, \hat{\sigma}) \\
&= \delta_w(p, \hat{\sigma}) \\
&= \delta_v(p, \sigma).
\end{aligned}
$$

Hence, $M_w^\downarrow = (Q, \Sigma, \delta_w^\downarrow, q_0, F) = M_v$. In other words, the identification of $M_v$ and $M_w$ leads to that the retraction of $M_v$ is identical with itself.

The next result shows that the concept of generalized extensions is a generalization of the extensions used in [12].

**Proposition 7.1.** *Let $M_v = (Q, \Sigma, \delta_v, q_0, F)$ be a PACV. Then the extension $\hat{M} = (Q, \mathcal{D}(\Sigma), \hat{\delta}_v, q_0, F)$ given by Definition 2.2 is the same as the generalized extension $M_w^\uparrow = (Q, \mathcal{D}(\Sigma), \delta_w^\uparrow, q_0, F)$.*

**Proof.** It is sufficient to show that $\hat{\delta}_v(p, W') = \delta_w^\uparrow(p, W')$ for all $(p, W') \in Q \times \mathcal{D}(\Sigma)$. By Definition 2.2, we see that $\hat{\delta}_v(p, W') = \sum_{\sigma \in \Sigma} W'(\sigma) \cdot \delta_v(p, \sigma)$. On the other hand, it follows from Definition 4.1 that

$$\begin{aligned}
\delta_w^\uparrow(p, W') &= \sum_{\hat{\tau} \in \Sigma_w} \left[ \sum_{\sigma \in \Sigma} W'(\sigma) \cdot \frac{\hat{\tau}(\sigma)}{\sum_{\hat{\pi} \in \Sigma_w} \hat{\pi}(\sigma)} \right] \cdot \delta_w(p, \hat{\tau}) \\
&= \sum_{\hat{\tau} \in \Sigma_w} \left[ \sum_{\sigma \in \Sigma} W'(\sigma) \cdot \hat{\tau}(\sigma) \right] \cdot \delta_w(p, \hat{\tau}) \\
&= \sum_{\hat{\tau} \in \Sigma_w} W'(\tau) \cdot \delta_w(p, \hat{\tau}) \\
&= \sum_{\tau \in \Sigma_w} W'(\tau) \cdot \delta_v(p, \tau).
\end{aligned}$$

Hence, $\hat{\delta}_v(p, W') = \delta_w^\uparrow(p, W')$, as desired. □

Based on Proposition 7.1, we view the extension in Definition 2.2 as a generalized extension hereafter. As we see from Figure 1, there are two approaches from computing with words to computing with all words: One is the generalized extension (*b*); the other is the composition of processes (*a*) and (*c*). The next proposition shows that both approaches yield the same result for the model of probabilistic automata.

**Proposition 7.2.** Let $M_w = (Q, \Sigma_w, \delta, q_0, F)$ be a PACW. Then $(M_w^\downarrow)^\uparrow = M_w^\uparrow$.

**Proof.** By definition, $M_w^\uparrow = (Q, \mathcal{D}(\Sigma), \delta^\uparrow, q_0, F)$ with $\delta^\uparrow(p, W') = \sum_{W \in \Sigma_w} \theta_{W'}(W) \cdot \delta(p, W)$ for any $(p, W') \in Q \times \mathcal{D}(\Sigma)$. In contrast, $M_w^\downarrow = (Q, \Sigma, \delta^\downarrow, q_0, F)$, where $\delta^\downarrow(p, \sigma) = \sum_{W \in \Sigma_w} \chi_\sigma(W) \cdot \delta(p, W)$ for any $(p, \sigma) \in Q \times \Sigma$. Consequently, the generalized extension of $M_w^\downarrow$ is $(M_w^\downarrow)^\uparrow = (Q, \mathcal{D}(\Sigma), (\delta^\downarrow)^\uparrow, q_0, F)$. By Proposition 7.1 and Definition 4.1, we see that

$$\begin{aligned}
(\delta^\downarrow)^\uparrow(p, W') &= \sum_{\sigma \in \Sigma} W'(\sigma) \cdot \delta^\downarrow(p, \sigma) \\
&= \sum_{\sigma \in \Sigma} W'(\sigma) \cdot \left[ \sum_{W \in \Sigma_w} \chi_\sigma(W) \cdot \delta(p, W) \right] \\
&= \sum_{\sigma \in \Sigma} \sum_{W \in \Sigma_w} W'(\sigma) \cdot \chi_\sigma(W) \cdot \delta(p, W) \\
&= \sum_{W \in \Sigma_w} \sum_{\sigma \in \Sigma} W'(\sigma) \cdot \chi_\sigma(W) \cdot \delta(p, W) \\
&= \sum_{W \in \Sigma_w} \left[ \sum_{\sigma \in \Sigma} W'(\sigma) \cdot \chi_\sigma(W) \right] \cdot \delta(p, W) \\
&= \sum_{W \in \Sigma_w} \theta_{W'}(W) \cdot \delta(p, W) \\
&= \delta^\uparrow(p, W')
\end{aligned}$$

for any $(p, W') \in Q \times \mathcal{D}(\Sigma)$. That is, $(\delta^\downarrow)^\uparrow = \delta^\uparrow$, and thus $(M_w^\downarrow)^\uparrow = M_w^\uparrow$, finishing the proof. □

**Remark 7.3.** Observe that the propositions obtained in this section evidently hold when the words are interpreted as possibility distributions. By using the properties of equivalences developed in Section 6, one can derive the results analogous to Propositions 7.1 and 7.2 for the model of probabilistic grammars.

## 8  Conclusion

In this paper, we have introduced two equivalent probabilistic models of computing with words via probabilistic automata and probabilistic grammars. The work has been developed based on the probabilistic distribution interpretation of words; some necessary modifications have been remarked in order that the results are applicable to the possibility distribution interpretation of words. The probabilistic models here can easily incorporate expert knowledge described by propositions in a natural language into system specification. Taking the finite inputs of the models into account, we established their retractions and generalized extensions, which have made it possible to compute any words. Furthermore, we obtained a retraction principle and a generalized extension principle showing that computing with values and computing with all words can be respectively implemented by computing with some special words. Moreover, the retractions and the generalized extensions proved to be equivalence-preserving. In addition, we have examined some analytical properties of transition probability functions of the generalized extensions, which are helpful in comparing the transition probabilities of two near inputs. Some relationships among the retractions, the generalized extensions, and the extensions studied recently in [12] have also been provided.

There are some limits and directions in which the present work can be extended. As mentioned earlier, the generalized extension of a probabilistic model is actually a process of interpolation. Thus, a basic problem is how to choose words and how to rationally specify their behavior. In turn, one can use many other interpolation approaches to cope with the problem of accepting any words as inputs. As a continuation of [2], this work further indicates that building a model for computing with some special words and then extending the model for computing with all words are of universality. Therefore, it is feasible to apply this method to other computational models such as fuzzy grammars [8], other probabilistic automata [16], and fuzzy and probabilistic neural networks (see, for example, [17, 1]). A topic of ongoing work concerns the formal model of computing with words of many kinds.

## Acknowledgment

The authors would like to thank Mehryar Mohri, Franck Thollard, and Enrique Vidal for their kind help on some references.

## References

[1]  J.J. Buckley, Y. Hayashi, Fuzzy neural networks: A survey, Fuzzy Sets and Systems, 66 (1994) 1-13.

[2]  Y.Z. Cao, M.S. Ying, G.Q. Chen, Retraction and generalized extensions of computing with words, available at http://arxiv.org/abs/cs.AI/0604070, April 2006.

[3]  K.S. Fu, Syntactic Pattern Recognition and Applications, Prince-Hall, Inc., Englewood Cliffs, New Jersey, 1982.

[4]  K.S. Fu, T. Li, On stochastic automata and languages, Inform. Sci. 1 (1969) 403-19.

[5]  D. Fudenberg, J. Tirole, Game Theory, MIT Press, Cambridge, MA, 1991.

[6] J.E. Hopcroft, J.D. Ullman, Introduction to Automata Theory, Languages, and Computation, Addison-Wesley, Reading, MA, 2001.

[7] R.I. John, P.R. Innocent, Modeling uncertainty in clinical diagnosis using fuzzy logic, IEEE Trans. Syst., Man, Cybern., Part B, 35 (2005) 1340-1350.

[8] A. Kandel, S.C. Lee, Fuzzy Switching and Automata: Theory and Applications, New York: Russak, 1979.

[9] M. Margaliot, G. Langholz, Fuzzy control of a benchmark problem: A computing with words approach, IEEE Trans. Fuzzy Syst. 12 (2004) 230-235.

[10] J.N. Mordeson, D.S. Malik, Fuzzy Automata and Languages: Theory and Applications, Boca Raton, FL: Chapman & Hall/CRC, 2002.

[11] A. Paz, Introduction to Probabilistic Automata, New York: Academic, 1971.

[12] D.W. Qiu, H.Q. Wang, A probabilistic model of computing with words, J. Comput. Syst. Sci. 70 (2005) 176-200.

[13] M.O. Rabin, Probabilistic automata, Inform. Contr. 6 (1963) 230-245.

[14] A. Salomaa, Probabilistic and weighted grammars, Inform. Contr. 15 (1969) 529-544.

[15] E.S. Santos, Maxmin automata, Inform. Contr. 13 (1968) 363-377.

[16] A. Sokolova, E.P. de Vink, Probabilistic automata: System types, parallel composition and comparison, in Validation of Stochastic Systems, LNCS 2925, 2004, pp. 1-43.

[17] D.F. Specht, Probabilistic neural networks for classification, mapping or associative memory, in Proc. IEEE Int. Conf. Neural Networks, vol. 1, 1988, pp. 525-532.

[18] S.F. Thomas, Fuzziness and Probability, ACG Press, Wichita, KS, 1995.

[19] H.Q. Wang, D.W. Qiu, Computing with words via Turing machines: A formal approach, IEEE Trans. Fuzzy Syst. 11 (2003) 742-753.

[20] C.S. Wetherell, Probabilistic languages: A review and some open questions, Comput. Surveys, 12 (1980) 361-379.

[21] R.R. Yager, Defending against strategic manipulation in uninorm-based multi-agent decision making, Eur. J. Oper. Res. 141 (2002) 217-232.

[22] M.S. Ying, A formal model of computing with words, IEEE Trans. Fuzzy Syst. 10 (2002) 640-652.

[23] L.A. Zadeh, Fuzzy logic, neural networks, and soft computing, Commun. ACM, 37 (1994) 77-84.

[24] ——, Probability theory and fuzzy logic are complementary rather than competitive, Technometrics, 37 (1995) 271-276.

[25] ——, Fuzzy Logic = computing with words, IEEE Trans. Fuzzy Syst. 4 (1996) 103-111.

[26] ——, From computing with numbers to computing with words—From manipulation of measurements to manipulation of perceptions, IEEE Trans. Circuits Syst. I: Fund. Theory Appl. 45 (1999) 105-119.

[27] ——, Outline of a computational theory of perceptions based on computing with words, in Soft Computing and Intelligent Systems, N. K. Sinha and M. M. Gupta, Eds. Boston, MA: Academic, 1999, pp. 3-22.

[28] ——, A new direction in AI: Toward a computational theory of perceptions, AI Mag. 22 (2001) 73-84.

[29] ——, Toward a perception-based theory of probabilistic reasoning with imprecise probabilities, J. Stat. Plan. Infer. 105 (2002) 233-264.

[30] L.A. Zadeh, J. Kacprzyk, Computing With Words in Information/Intelligent Systems, Heidelberg, Germany: Physica-Verlag, vol. 1 and vol. 2, 1999.